\documentclass[lettersize,journal]{IEEEtran}
\usepackage{amsmath,amsfonts}
\usepackage{textcomp}
\usepackage{stfloats}
\usepackage{verbatim}
\usepackage{cite}
\usepackage{subcaption,float}
\usepackage[pdftex]{graphicx}
\usepackage{color, colortbl}
\usepackage{amsmath}
\usepackage{multirow, array}
\usepackage{bm,url}
\usepackage{breakurl}
\usepackage{enumerate}
\usepackage[table,xcdraw]{xcolor}
\usepackage{hhline}
\usepackage{rotating}
\usepackage{adjustbox}
\usepackage{tikz}
\usepackage{collcell}
\usepackage[ruled,vlined]{algorithm2e}

\usepackage{pgfplots}

\SetKwInput{KwInput}{Input}  
\SetKwInput{KwOutput}{Output} 

\renewcommand{\baselinestretch}{1.0}

\begin{document}

\title{Constrained Twin Variational Auto-Encoder for Intrusion Detection in IoT Systems}

\author{
	\IEEEauthorblockN{
		Phai~Vu~Dinh,
		Quang~Uy~Nguyen,
		Dinh~Thai~Hoang,
		Diep~N.~Nguyen,  \\
		Son~Pham~Bao, and
		Eryk~Dutkiewicz
	}\\
	
	\thanks{
    V. D. Phai, D. T. Hoang, Diep N. Nguyen, and E. Dutkiewicz are with the School of Electrical and Data Engineering, the University of Technology Sydney, Sydney, NSW 2007, Australia (e-mail:   Phai.D.Vu@student.uts.edu.au, \{hoang.dinh, diep.nguyen, eryk.dutkiewicz\}@uts.edu.au). \\
    N. Q. Uy is with the Computer Science Department, Faculty of Information Technology, Le Quy Don Technical University, Hanoi, Vietnam (email: quanguyhn@lqdtu.edu.vn). \\
    P. B. Son is with the University of Engineering and Technology, Vietnam National University, Hanoi, Vietnam (e-mail: sonpb@vnu.edu.vn). \\
	Corresponding author: Quang Uy Nguyen (email: quanguyhn@lqdtu.edu.vn)
 }	
}



\maketitle

\begin{abstract}
Intrusion detection systems (IDSs) play a critical role in protecting billions of IoT devices from malicious attacks. However, the IDSs for IoT devices face inherent challenges of IoT systems, including the heterogeneity of IoT data/devices, the high dimensionality of training data, and the imbalanced data. Moreover, the deployment of IDSs on IoT systems is challenging, and sometimes impossible, due to the limited resources such as memory/storage and computing capability of typical IoT devices. To tackle these challenges, this article proposes a novel deep neural network/architecture called Constrained Twin Variational Auto-Encoder (CTVAE) that can feed classifiers of IDSs with more separable/distinguishable and lower-dimensional representation data. Additionally, in comparison to the state-of-the-art neural networks used in IDSs, CTVAE requires less memory/storage and computing power, hence making it more suitable for IoT IDS systems. Extensive experiments with the 11 most popular IoT botnet datasets show that CTVAE can boost around 1\% in terms of accuracy and Fscore in detection attack compared to the state-of-the-art machine learning and representation learning methods, whilst the running time for attack detection is lower than $2E{\mbox{-}}6$ seconds and the model size is lower than 1 MB. We also further investigate various characteristics of CTVAE in the latent space and in the reconstruction representation to demonstrate its efficacy compared with current well-known methods. 
\end{abstract}


\section{Introduction}
\IEEEPARstart{I}{ntrusion} Detection Systems (IDSs) are one of the most effective solutions to secure billions of IoT devices from malicious attacks \cite{R11}. However, IDSs face various difficulties and challenges posed by IoT networks. There are four main challenges of IDSs when applied to IoT systems. The first challenge is complex data collected from IoT devices that are often of high dimension with heterogeneous data formats. Specifically, given the heterogeneity of IoT devices, the data collected from them often contain highly-complex structures, e.g., correlated features, redundant data, and/or high dimensions \cite{R7, R4}. For example, data can be collected from network traffic, system calls, and user activity logs  of different IoT devices. The collection of data from both network traffic and host devices increases the number of features and further complicates the relationship among different features. In addition, the daily evolution of malicious software is another obstacle since the IDSs can be bypassed by unknown attacks. Second, the introduction of new IoT device types and their manufacturers also makes it difficult to properly and effectively update/patch existing IDSs. Attackers can discover new vulnerabilities to make zero-day attacks on IoT devices \cite{R8, R9}. Moreover, the collection of attack data is often difficult compared to the process of collecting benign data, leading to an imbalanced dataset. The third challenge also comes from zero-day attacks, leading to a failure of the engine detection due to the detection models not being trained by data samples of the zero-day attacks. The fourth challenge is the deployment of IDSs on IoT devices that are often limited in energy, computing, storage/memory, and communications resources. These make advanced IDSs that require intensive computing capability and are demanding in storage/memory/energy inapplicable to IoT devices.

One can categorize IDSs into three types, signature-based, anomaly-based, and hybrid
\cite{hajiheidari2019intrusion}. The signature-based IDSs attempt to detect an attack by matching features of the current network data with pre-defined attack patterns. IDSs using this method can quickly detect well-known attacks and they can be used at the earliest stage in protecting IoT systems. However, the signature-based IDSs face challenges posed by new attacks with  unknown patterns. The anomaly-based IDSs are designed to learn from the known patterns to detect the new attacks of which their characteristics are unknown in the pre-defined patterns. The hybrid IDSs are a combination of both signature-based and anomaly-based IDSs. Consequently, the hybrid IDSs can gain higher accuracy and less False Alarm Rate (FAR) in detecting the anomaly IoT attacks, at the cost of higher computation and resources of IoT systems.

Machine learning (ML) has recently found its numerous applications in IDSs for IoT systems, making them more robust and effective in detecting incidents. However, ML algorithms face challenges from the complexity of IoT data, e.g., correlated features and high dimensions \cite{ids2019survey}, \cite{ferrag2020deep}. The former links to the relationship of two or more features of datasets, whilst the latter means that the number of features for each data sample is staggeringly high. To address these challenges, many representation learning (RL) methods have been proposed, and they play a pivotal role in the success of ML in IDSs \cite{bengio2013rep}. Effective RL algorithms  make input data more separable for classifiers. Moreover, these representation algorithms can provide classifiers with a more abstract representation of the data \cite{Goodfellow} by mapping the original data into new representations to find the most important information, i.e., patterns, and anomalies. Amongst many RL methods, Auto-Encoders (AEs) are one of the most popular models due to their effectiveness and convenience \cite{R26}. An AE aims to learn the best parameters required to reconstruct its input at the output \cite{dao2021stacked}. For that, AEs are very helpful in RL by transforming the original representation into a new representation at the bottleneck layers of AEs. 
However, AEs cannot create a desired data distribution at its bottleneck that is critical to precisely control the latent vector to minimize information redundancy for large-scale datasets, often encountered by IoT IDSs \cite{R27}. A variant of AE, the Variational Auto-encoder (VAE) can approximate a distribution at
the output of its encoder that provides a significant control/practice over its latent space by using prior distribution. Nevertheless, the performance of the latent representation of the original VAE is often worse than those of AEs~\cite{R1, TomczakW17}. The reason is that the latent representation of VAE is sarcastically sampled from a distribution, and this representation is not stable to use as the input of a classifier. In addition, the performance of the latent representation of VAE might not be effective with multi-classes datasets since VAE's latent vector is sampled from a normal distribution.

To improve the effectiveness of the data representation of VAE, we first propose a new model called Twin Variational Auto-Encoders (TVAE). TVAE learns to convert the stochastic representation of the latent space into a deterministic representation at its output. Our preliminary results in \cite{R20} show the superior performance of TVAE compared with several state-of-the-art RL methods. However, this early TVAE model was trained in an unsupervised mode and ignored the label information of datasets. To leverage the label information of datasets to force its latent vector of different classes into different Gaussian distributions, we then design a novel deep neural network called Constrained TVAE (CTVAE). Therefore, data samples of different classes in the latent space are directly separated. Note that, the output representation of CTVAE is to reconstruct its latent representation, thus this representation is called the \textit{reconstruction representation}. It is worth noting that CTVAE can address the challenge of high dimensions by transforming IoT IDS input data having high dimensions into a new representation that has lower dimensions, to obtain more abstract representation data. CTVAE also combines with K-Mean and silhouette values \cite{R29} to find reasonable distributions of the major classes, introducing a solution to address the imbalanced data in IoT IDS systems. In addition, CTVAE can generate different Gaussian distributions in which data samples of different classes are separated.  Therefore, the data representation of CTVAE makes it easier for the classifiers to distinguish between normal and known attack data for the training set, and then enhances and facilitates the IDS classifiers in distinguishing between normal data and anomalies for the testing set.  Finally, CTVAE requires less storage/memory and computing power and hence can be more suitable for IoT IDS systems. Although the CTVAE is trained with three parts, e.g., an Encoder, a Hermaphrodite, and a Decoder, we only use the Decoder to extract the data representation. As a result, the running time and model size of the CTVAE are more suitable for IoT devices. The experiments performed on 11 IoT botnet datasets \cite{R13} show the superior performance in terms of accuracy and Fscore of the CTVAE over AE and VAE variants \cite{R1, R10, R13, R16, R27, R18, R19}, Xgboost \cite{R23}, Convolutional Sparse Auto-Encoder (CSAE) \cite{R18}. Our major contributions are summarized as follows:
		\begin{itemize}
			\item We propose a novel neural network model, i.e., CTVAE. CTVAE is trained in a supervised manner to construct the reconstruction representation that is more separable for different classes. 
			\item We conduct intensive experiments using 11 IoT datasets to evaluate the performance of CTVAE compared to other methods. The experimental results show that CTVAE helps classifiers perform much better when compared to learning from the latent representations produced by the other AEs and VAEs. The performance of CTVAE in terms of accuracy and Fscore is also better than recent state-of-the-art machine learning methods whilst the CTVAE model requires less storage/memory and computing power for IoT IDS systems.
			\item We leverage CTVAE to address the challenge of the imbalanced dataset. CTVAE can also support to detection of abnormal data.
			\item We also study various characteristics of the latent representation and the reconstruction representation  of CTVAE to justify its superior performance over other methods. 
		\end{itemize}

The remainder of this paper is organized as follows. Section \ref{related_work} discusses related work and Section \ref{back_ground} provides relevant backgrounds. In Section \ref{proposed_method}, we describe our proposed model, i.e., CTVAE. Section \ref{experiment_setting} presents the experimental settings. Discussion and analysis of our models are presented in Section \ref{result_analysis}. Finally, conclusions are drawn in Section \ref{conclusion}.
\section{Related work}	
\label{related_work}
This section reviews recent methods and the state-of-the-art RL models based on AEs and VAEs which are applied to IDSs. 

\subsection{Supervised Machine Learning Methods for IDSs}
ML has been widely used to develop detection models for IDSs. Depending the way of the training method, ML can be classified into three groups, e.g., supervised, unsupervised, and semi-supervised learning. The supervised methods use the label information to train a detection model. In contrast, the unsupervised methods build a detection model from data without the information of labels. These models are often used for discovering a novel/anomaly attack. The semi-supervised modes are a combination of both supervised and unsupervised methods.
In this subsection, we present prevalent supervised ML methods used to develop models for IDSs.	The analysed models include Logistic Regression (LR) \cite{kleinbaum2002logistic}, Support Vector Machine (SVM) \cite{LSVM}, Decision Tree (DT)\cite{DT}, Random Forests (RF) \cite{RF}, and Xgboost (Xgb) \cite{R23}. 

LR is a popular ML algorithm that uses statistical methods based on prior observation of the input data to make a prediction of a binary result, e.g., benign traffic or anomaly decisions. In spite of showing up in the early stage of ML eras, LR is widely applied to build IDSs nowadays due to its effectiveness in terms of accuracy and low cost of computation. Similarly, in this stage, DT is another popular classifier used to build the IDSs. DT attempts to create a tree of decisions in which its leaves can present a target value as the normal or malicious data~\cite{sindhu2012DT,moon2017DT}. The SVM model  was presented in the early 2000's. It generates the hyper-plane in the feature space to distinguish the normal with the attack data ~\cite{chen2009SVM, hasan2014SVM}. In the 2010s, the RF classifier gained state-of-the-art results in many domains including IDSs. The RF model is developed from trees of decisions which uses an ensemble learning method to make a classification. RF model has achieved good performance in IDSs as shown in~\cite{li2020AERF,kim2006RF}. After the success of RF, Xgb was introduced by Chen in 2016~\cite{chen2016xgb}. The Xgb achieves state-of-the-art results on many ML challenges, and it is considered to be a convenient way to improve the efficiency and accuracy of IDSs~\cite{bhati2021Xgb}.

Although the above ML classifiers have successfully found their applications in network IDSs, for the IoT IDSs these classifiers face challenges from the complexity of  IoT systems. To tackle them, we present a new approach that uses the RL technique to map the IoT input data into a new space. This technique can facilitate ML models in improving the performance of IDSs in IoT systems.

\subsection{Representation Learning Methods for IDSs}	
	\begin{table*}[htp]
		\centering
		\setlength\tabcolsep{3pt}
		\caption{Differences between previous models and CTVAE.}
		\label{tab:ctvae-compare-previous1}
		\begin{tabular}{|l|l|l|}
			\hline
			\textbf{Criterion}& \textbf{\begin{tabular}{l}
					Previous ones, i.e., AE and VAE \\
					variants \cite{R1, R10, R13, R16, R27, R18, R19}
			\end{tabular}} & \textbf{Proposed model, i.e., CTVAE}   \\ \hline
			Architecture  & \begin{tabular}[c]{@{}l@{}}Use the architecture of the AE including  an Encoder  and a \\Decoder.\end{tabular} & \begin{tabular}[c]{@{}l@{}}We propose a novel architecture with three components,i.e., an Encoder,\\ a Hermaphrodite and a Decoder, instead of two components of AE \\variants.\end{tabular} \\ \hline
			\begin{tabular}[c]{@{}l@{}} \\ Extracting data\end{tabular} & \begin{tabular}[c]{@{}l@{}}Extract data from latent space of the  output of the Encoder.\end{tabular} & \begin{tabular}[c]{@{}l@{}}Extract data from the Decoder of CTVAE.\end{tabular}\\ \hline
			\begin{tabular}[c]{@{}l@{}}Technique used\\ \end{tabular} & \begin{tabular}[c]{@{}l@{}}Use penalized terms to force data samples in latent space \\during the process of training. The  problem of this method\\ is  a trade-off between the reconstruction term and the \\penalized term. This lowers the accuracy of classifiers when \\ using the data representation of these models.\end{tabular} & \begin{tabular}[c]{@{}l@{}}Separate data samples of different classes in the latent space by using \\separated Gaussian distributions. The decoder of the CTVAE attempts\\ to learn from data samples of the separated Gaussian distributions. \\\end{tabular} \\ \hline
			Effectiveness & \begin{tabular}[c]{@{}l@{}}AE variants may lower accuracy and Fscore for classifiers \\using their data representation. It also may not be effective\\ for multi-class datasets, as observed in Table \ref{tab:tab_multi_label_TVAE_vs_ALL} of \\ the paper .\end{tabular}& \begin{tabular}[c]{@{}l@{}}CTVAE increases accuracy and Fscore for classifiers using its data \\representation for both binary and multi-class datasets purpose,\\ as observed in Table \ref{tab:tab_multi_label_TVAE_vs_ALL} and \ref{tab:tab_result_anomaly_detection} in the paper.\end{tabular} \\ \hline
		\end{tabular}
		
	\end{table*}

This subsection reviews some RL methods that are used to facilitate IDSs. Most of the RL methods rely on the architecture of AEs \cite{vu2020deep, vincent2010stacked, Shone2018, R19}. Vincent and Hinton et al.~\cite{vincent2010stacked}, \cite{hinton2006reducing} used AE as a non-linear transformation to discover unknown data structures compared to those of Principal Component Analysis (PCA). Shone et al.~\cite{Shone2018} proposed a non-symmetric deep-auto encoder (NDAE) which only uses an encoder for both encoded and decoded tasks. This architecture can extract data from the latent space to enhance the performance of Random Forest (RF). Similarly, the authors in \cite{R19} introduced a self-taught learning based on combining sparse auto-encoder with Support Vector Machine (SAE–SVM). Both NDAE and SAE-SVM were shown to yield higher accuracy in IDSs, compared with the previous works. However, it is unclear whether the improvement comes from the latent space or the advantage of RF and SVM.  


Recently, Luo et al.~\cite{R18} proposed Convolutional Sparse Auto-Encoder (CSAE), which leverages the structure of the convolutional AE and incorporates the max-pooling to heuristically sparsify the feature maps for RL. The encoder of the CSAE is used as a pre-trained model for RL. After training, the encoder of CSAE is also added by a softmax layer and this network is trained in a supervised mode. The CSAE is then considered as a supervised learning model for RL and this configuration is referred to as CSAEC~\cite{phan2018Tree}. VAE provides significant control over the latent vector to minimize information redundancy with large-scale datasets \cite{R27}. The VAE model also generates data examples that improve the performance of deep-learning models in handling imbalanced datasets \cite{Abdulhammed2018}. However, VAE is hardly used for RL as its latent representation is stochastically sampled from a Gaussian distribution thus it is unstable when used as the input to classification algorithms \cite{jang2017}. To achieve a more robust representation in VAE, the authors in \cite{bVAE} and \cite{wu2020vector} added an extra hyperparameter $\beta$ to the VAE and a quantization term in the bottleneck of VAE to introduce $\beta$-VAE and VQVAE (Vector Quantized VAE) models, respectively.

More recently, label information has been used to enhance the representation based on  AEs. Vu et al.~\cite{R1} introduced Multi-distribution AE (MAE) and Muti-distribution Variational Auto-Encoder (MVAE) for learning latent representation.  MAE and MVAE used label information in datasets to project original data into tightly separated regions of the latent space by adding a penalized component in the loss function. Although MAE and MVAE have proven to be superior in binary datasets, the process of projecting data may not be effective with multiclass datasets. For the multi-class datasets, the latent space of MAE and MVAE can be irregular due to the means of the distribution in the latent space being fixed on a hyperplane. To highlight the
fundamental differences and the knowledge gaps of CTVAE in comparison with previous articles, we present in Table \ref{tab:ctvae-compare-previous1}.

Given the above, this paper proposes two novel deep learning (DL) models: TVAE and CTVAE. The TVAE can reconstruct the latent representation to become the reconstruction representation and use this reconstruction representation as a new representation for classifiers. The TVAE does not require label information for the training process, whilst the CTVAE can use the label information to force the latent vector of classes to different separable Gaussian distributions. Therefore, the decoder of CTVAE can effectively separate the Gaussian distributions in the reconstruction space, making this representation to be easier for classifiers. 

\section{Background}
\label{back_ground}
\begin{figure}[t]
	\centering
	\includegraphics[width=0.7\linewidth]
	{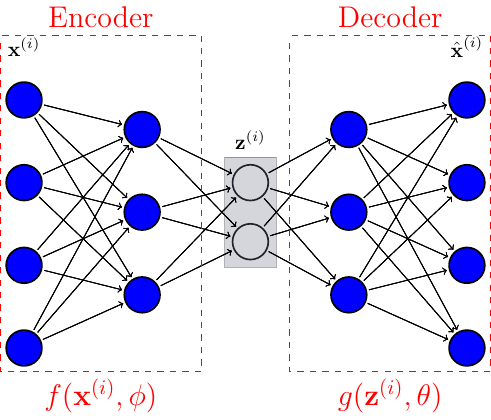}
	\caption{Auto-Encoder (AE).}
	\label{fig:ae_fig} 
\end{figure}
\begin{figure}[t]
	\centering
	\includegraphics[width=0.9\linewidth]
	{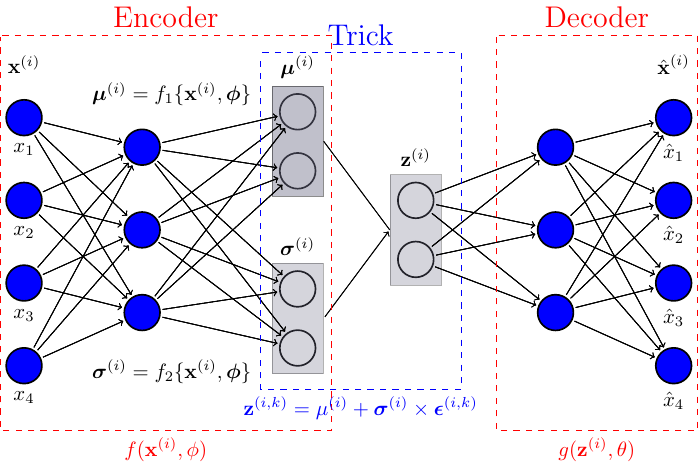}
	\caption{Variational Auto-Encoder (VAE).}
	\label{fig:vae_fig} 
\end{figure}
This section presents AE and VAE in detail. They are the two key components of the proposed models in the next section.  
\subsection{Auto-Encoder (AE)}
An Auto-Encoder is a neural network trained by unsupervised learning to learn the best parameters required to reconstruct its output as close to its input as possible \cite{R26}. An AE has two parts, an encoder, and a decoder, as illustrated in Fig. \ref{fig:ae_fig}. Let ${\textbf{W}}$, ${\textbf{b}}$, ${\textbf{W}^{'}}$, and ${\textbf{b}^{'}}$ be the weight matrix and biases of the encoder and decoder, respectively. $\boldsymbol{\phi}$ = (${\textbf{W}}$, ${\textbf{b}}$) is the parameter sets of the encoder function ${\textbf{z}}$ = ${f_{\boldsymbol{\phi}}(\textbf{x})}$, while ${\boldsymbol{\theta}}$ = (${\textbf{W}^{'}}$, ${\textbf{b}^{'}}$) is the parameter sets of the decoder function ${\hat{\textbf{x}}}$ = ${g_{\boldsymbol{\theta}}(\textbf{z})}$. We use a dataset ${\textbf{X}} = \{\textbf{x}^{(1)}, \textbf{x}^{(2)},..,\textbf{x}^{(n)}\} = \{\textbf{x}^{(i)}\}_{i=1}^{n}$ to train the AE model which has its loss function as follows:
\begin{equation}
\label{eq:ae_loss}
\ell_{\mbox{AE}}(\textbf{X}, \boldsymbol{\phi}, \boldsymbol{\theta})=\frac{1}{n} \sum_{i=1}^{n}\left(\textbf{x}^{(i)}-\hat{\textbf{x}}^{(i)}\right)^{2},
\end{equation}
where $\hat{\textbf{x}}^{(i)}$ is the output of the AE corresponding with the input $\textbf{x}^{(i)}$. The latent representation of the encoder is usually referred to as a bottleneck which is used as an input for classifiers.
\subsection{Variational Auto-Encoder (VAE)}	
A Variational Auto-Encoder (Fig. \ref{fig:vae_fig}) is another version of AE~\cite{Kingma_2019}. The difference between a VAE and an AE is that the latent representation ($\textbf{z}$) of the VAE is sampled from a Gaussian distribution parameterized by mean ($\boldsymbol{\mu}$) and standard deviation ($\boldsymbol{\sigma}$). We define the encoder by $q_{\boldsymbol{\phi}}(\textbf{z}^{}|\textbf{x}^{})$ and the decoder by $p_{\boldsymbol{\theta}}(\textbf{x}^{}|\textbf{z}^{})$, then the loss function of a VAE for a data point $\textbf{x}^{(i)}$ includes two terms as follows:
\begin{equation}
\label{eq:loss_vae}
\begin{aligned}
{\ell_{\mbox{VAE}}} (\textbf{x}^{(i)}, \boldsymbol{\theta}, \boldsymbol{\phi})
=D_{\mbox{KL}}\big(q_{\boldsymbol{\phi}}(\textbf{z}^{}|\textbf{x}^{(i)})| p_{\boldsymbol{\theta}}(\textbf{z}^{})\big) \\  - E_{q_{\boldsymbol{\phi}}(\textbf{z}^{}|\textbf{x}^{(i)})}[\log p_{\boldsymbol{\theta}}(\textbf{x}^{(i)}|\textbf{z}^{})] \cdot
\end{aligned}
\end{equation}

The first term in Eq. (\ref{eq:loss_vae}) is the KL divergence between the approximation posterior ($q_{\boldsymbol{\phi}}(\textbf{z}^{}|\textbf{x}^{(i)})$) and the prior distribution ($p_{\boldsymbol{\theta}}(\textbf{z}^{})$). This divergence measures how close the posterior is to the prior. The second term $-E_{q_{\boldsymbol{\phi}}(\textbf{z}^{}|\textbf{x}^{(i)})}[\log p_{\boldsymbol{\theta}}(\textbf{x}^{(i)}|\textbf{z}^{})]$ is reconstruction error (RE) of the VAE. This term forces the decoder to learn to reconstruct the input data.

Since it is tricky to generate a sample $\textbf{z}^{(i)}$ from $q_{\boldsymbol{\phi}}(\textbf{z}^{}|\textbf{x}^{})$, a reparameterization trick in Eq. (\ref{eq:sampling_trick_vae}) is used to overcome the high variance problem when applying the Monte Carlo method \cite{An2015VariationalAB}. In particular, instead of using a random variable from the original distribution, the reparameterization trick uses a random variable $\textbf{z}^{(i)}$ from a standard normal distribution as follows:
\begin{equation}
\label{eq:sampling_trick_vae}
\begin{aligned}
{\textbf{z}}^{(i,k)}= \boldsymbol{\mu}^{(i)} + \boldsymbol{\sigma}^{(i)} \times \boldsymbol{\epsilon}^{(i,k)}; ~~~ \boldsymbol{\epsilon}^{(i,k)}\sim N(0,I),
\end{aligned}
\end{equation} 
where $\boldsymbol{\mu}^{(i)}$ and $\boldsymbol{\sigma}^{(i)}$ are the mean and standard deviation of the Gaussian distribution of an individual latent variable $\textbf{z}^{(i)}$, respectively. The values of $\boldsymbol{\mu}^{(i)}$ and $\boldsymbol{\sigma}^{(i)}$ are obtained via the encoder by using functions $\boldsymbol{\mu}^{(i)}=f_1(\textbf{x}^i, \boldsymbol{\phi})$ and $\boldsymbol{\sigma}^{(i)}=f_2(\textbf{x}^{(i)}, \boldsymbol{\phi})$ as illustrated in Fig. \ref{fig:vae_fig}, respectively.

\section{Proposed Methodology}
\label{proposed_method}
This section presents both a novel deep-learning neural
network architecture/model: Twin Variational Auto-Encoder (TVAE) and a novel neural network model, i.e., Constrained TVAE (CTVAE).
\subsection{Twin Variational Auto-Encoder (TVAE)}
\begin{figure}[t]
	\centering
	\includegraphics[width=0.95\linewidth]
	{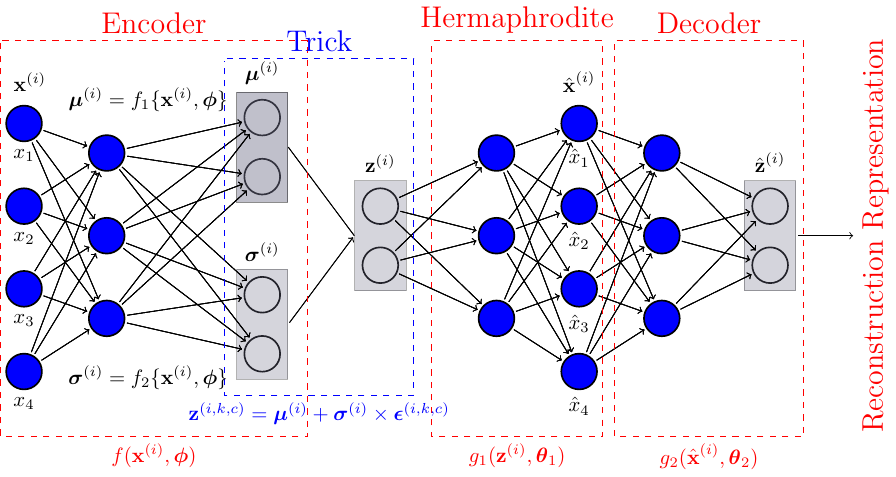}
	\caption{Twin Variational Auto-Encoder (TVAE) architecture.}
	\label{fig:tvae_fig} 
\end{figure}
TVAE was first proposed in our previous work~\cite{R20} for RL in an unsupervised manner. The architecture of TVAE is shown in Fig. \ref{fig:tvae_fig}. The TVAE has three components: an encoder, a hermaphrodite, and a decoder. Let $\boldsymbol{\phi}$, $\boldsymbol{\theta}_{1}$, $\boldsymbol{\theta}_{2}$ be the parameter sets of the encoder, the hermaphrodite, and the decoder, respectively. First, the encoder attempts to map an input sample $\textbf{x}^{(i)}$ to two hidden vectors $\boldsymbol{\mu}^{(i)}$ and $\boldsymbol{\sigma}^{(i)}$ by using functions $f_{1}(\textbf{x}^{(i)},\boldsymbol{\phi})$ and $f_{2}(\textbf{x}^{(i)},\boldsymbol{\phi})$, respectively. A feasible reparameterization trick is applied by using a deterministic function $\textbf{z}^{(i,k)}=\boldsymbol{\mu}^{(i)} +\boldsymbol{\sigma}^{(i)}  \times \boldsymbol{\epsilon}^{(i,k)}$, which helps to transfer the input $\textbf{x}^{(i)}$ into a new latent space $\textbf{z}^{(i)}$. Second, the hermaphrodite performed by function $g_{1}(\textbf{z}^{(i)}, \boldsymbol{\theta}_{1})$  has two roles. On the one hand, the hermaphrodite is the decoder of the VAE in TVAE. On the other hand, the hermaphrodite is the encoder of the AE. The third component in TVAE is the decoder  to reconstruct $\textbf{z}^{(i)}$ from  $\hat{\textbf{x}}^{(i)}$ by the function $g_{2}(\hat{\textbf{x}}^{(i)}, \boldsymbol{\theta}_{2})$. After training the TVAE model with dataset $\{\textbf{x}^{(i)}\}_{i=1}^n$, we input a data sample $\textbf{x}^{(i)}$ to the decoder to obtain its reconstruction representation $\hat{\textbf{z}}^{(i)}$. That has the same dimensionality as that of the latent variable $\textbf{z}^{(i)}$.

The loss function of TVAE for data sample $\textbf{x}^{(i)}$, $\ell_{\mbox{TVAE}}$, includes three terms in Eq. (\ref{eq:t12}). 
\begin{equation}
\label{eq:t12}
\begin{aligned}	
\ell_{\mbox{TVAE}} (\textbf{x}^{(i)}, \boldsymbol{\phi}, \boldsymbol{\theta}_{1}, \boldsymbol{\theta}_{2})
=  D_{\mbox{KL}}[q(\textbf{z}|\textbf{x}^{(i)})||p(\textbf{z})] \\-E_{q(\textbf{z}|\textbf{x}^{(i)})}\log p(\textbf{x}^{(i)}|\textbf{z}) - E_{q(\textbf{z}|\textbf{t}^{(i)})}\log p(\textbf{t}^{(i)}|\textbf{z}) \cdot
\end{aligned}
\end{equation}
The first term in Eq. (\ref{eq:t12}) is the KL divergence between the approximation posterior $q(\textbf{z}|\textbf{x}^{(i)})$ and the prior distribution $p(\textbf{z})$. The second term is the reconstruction error of the VAE and the third term is the reconstruction error of the AE. Assume that both target variables $\hat{\textbf{x}}^{(i)}$ and $\textbf{t}^{(i)}=\hat{\textbf{z}}^{(i)}$ are given by deterministic functions with additive Gaussian noise. The values of $-E_{q(\textbf{z}|\textbf{x}^{})}\log p(\textbf{x}^{}|\textbf{z})$ and $-E_{q(\textbf{z}|\textbf{t}^{})}\log p(\textbf{t}^{}|\textbf{z})$ are thus scaled with mean-squared-error function \cite{Bishop2006}, as follows:
\begin{equation}
\label{eq:t13}
\begin{aligned}	
-E_{q(\textbf{z}|\textbf{x})}\log p(\textbf{x}|\textbf{z})= \frac{1}{n} \sum_{i=1}^{n}(\textbf{x}^{(i)}-\hat{\textbf{x}}^{(i)})^{2}, \\
-E_{q(\textbf{z}|\textbf{t})}\log p(\textbf{t}|\textbf{z}) = \frac{1}{n} \sum_{i=1}^{n}(\textbf{z}^{(i)}-\hat{\textbf{z}}^{(i)})^{2} \cdot
\end{aligned}
\end{equation}
The KL divergence $D_{\mbox{KL}}[q(\textbf{z}|\textbf{x}^{(i)})||p(\textbf{z})]$ between the approximation posterior $q(\textbf{z}|\textbf{x}^{(i)})$ and the prior distribution $p(\textbf{z}^{})$ of the latent variable $\textbf{z}^{(i)}$ can be measured by assuming both of them being normal Gaussian distribution. Thus, the $\ell_{\mbox{TVAE}}$ for the dataset $\textbf{X}$ can be written as:
\begin{equation}
\label{eq:loss_tvae}
\begin{aligned}	
\ell_{\mbox{TVAE}} (\textbf{X}, \boldsymbol{\phi}, \boldsymbol{\theta}_{1}, \boldsymbol{\theta}_{2} )=\frac{1}{n} \sum_{i=1}^{n} \big{[} (\textbf{x}^{(i)}-\hat{\textbf{x}}^{(i)})^{2} + \beta_1 (\textbf{z}^{(i)}-\hat{\textbf{z}}^{(i)})^{2} \\ +   \frac{\beta_2}{2} \sum_{j=1}^{J}\big(-1-\log( ({\sigma}^{(i)}_{j})^{2})+({\mu}^{(i)}_{j})^{2}+({\sigma}^{(i)}_{j})^{2}\big) \big{]},
\end{aligned}
\end{equation}
where $\beta_1$ and $\beta_2$ are hyper-parameter settings to control the trade-off amongst three terms in (\ref{eq:loss_tvae}), $J$ denotes the dimension of the reconstruction representation, i.e., $\hat{\textbf{z}}^{(i)}$.
\subsection{Constrained Twin Variational Auto-Encoder (CTVAE)}
\begin{figure}[t]
	\centering
	\includegraphics[width=0.9\linewidth]
	{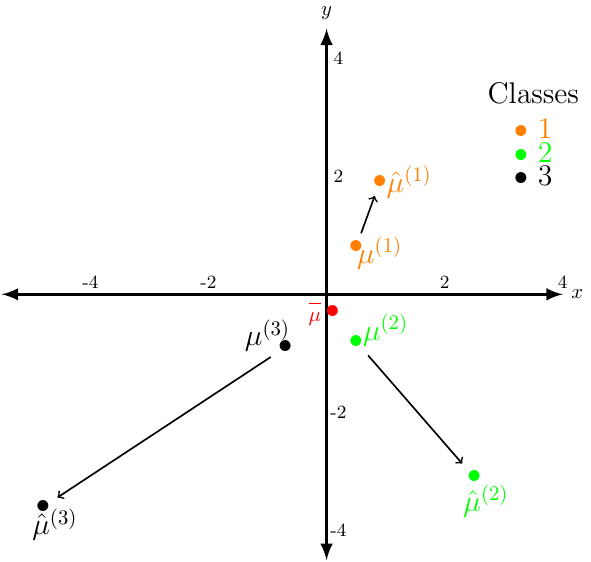}
	\caption{Moving $\boldsymbol{\mu}^{(c)}$ to $\hat{\boldsymbol{\mu}}^{(c)}$ of four classes.}
	\label{fig:algoScaledLatentSpace} 
\end{figure}
CTVAE is a novel supervised model developed from the unsupervised TVAE model. The idea is to use the label information to transform the latent representation ($\textbf{z}^{(i)}$) of each class into a distinguishable Gaussian distribution. After that, the decoder of CTVAE can learn to copy the distribution of $\textbf{z}^{(i)}$ into the reconstruction representation $\hat{\textbf{z}}^{(i)}$. Thus, the reconstruction representation $\hat{\textbf{z}}^{(i)}$ will be  more separated for different classes. To obtain this objective, the value of $\boldsymbol{\epsilon}^{(i,k)}$ is sampled from a distribution $N(\hat{\boldsymbol{\mu}}^{(c)}, \boldsymbol{\sigma}^{(c)})$ for each class $c$, instead of sampling from the same standard normal distribution $N(0, I)$. The reparameterization trick is rewritten as follows:
\begin{equation}
\label{eq:sampling_trick_tvaec}
\begin{aligned}
{\textbf{z}}^{(i,k,c)}= \boldsymbol{\mu}^{(i)} + \boldsymbol{\sigma}^{(i)} \times \boldsymbol{\epsilon}^{(i,k,c)}, \\  \space \boldsymbol{\epsilon}^{(i,k,c)}  \sim N(\hat{\boldsymbol{\mu}}^{(c)}, \boldsymbol{\sigma}^{(c)}) ,
\end{aligned}
\end{equation} 
where $\hat{\boldsymbol{\mu}}^{(c)}$ and $\boldsymbol{\sigma}^{(c)}$ are mean and standard deviation that are predefined for each class $c \in C$, respectively. Note that, the information of label $c$ is used for sampling $\boldsymbol{\epsilon}^{(i,k,c)}$ in Eq. (\ref{eq:sampling_trick_tvaec}).

		\begin{algorithm}[t]
			\SetAlgoLined
			\DontPrintSemicolon
			\KwInput{Dataset $\textbf{X}={\{\textbf{x}^{(i,c)}\}_{i=1}^n}$ where $\textbf{x}^{(i,c)}$ is $i^{th}$ sample in training set. $C$ is a set of classes. $c \in C$ is the class label. $|C|$ is the number of classes (i.e. $C=\{0,1,2\}, |C| = 3$). $d_z$ is the dimensionality of $\textbf{z}$, and $n_c$ and $n$ are the numbers of samples of label $c$ and training set, respectively.}
			\KwOutput{$\boldsymbol{\sigma}^{(c)}$, $\boldsymbol{\mu}^{(c)}$, and $\hat{\boldsymbol{\mu}}^{(c)}$ for each class $c$.}
			\textbf{Step 1}: Apply PCA to reduce data dimensionality.\;
			$\textbf{x}_{r}^{(i,c)}  = \mbox{PCA}(\textbf{x}^{(i,c)}, d_z)$\;
			\textbf{Step 2}: Calculate $\boldsymbol{\sigma}^{(c)}$, $\boldsymbol{\mu}^{(c)}$, and $\overline{\boldsymbol{\mu}}$ for each class.\;
			\ForEach {$c$ in $C$}{
				$\boldsymbol{\mu}^{(c)} = \frac{1}{n_c} \sum_{(i=1)}^{n_c} (\textbf{x}^{(i,c)}_{r})$\; $\boldsymbol{\sigma}^{(c)}=\sqrt{\frac{\sum_{(i=1)}^{n_c}|\textbf{x}^{(i,c)}_{r}-\boldsymbol{\mu}^{(c)}|^2}{n_c}}$ \;
			} 
			$\overline{\boldsymbol{\mu}}=\frac{1}{|C|} \sum_{c=1}^{|C|}\boldsymbol{\mu}^{(c)}$ \;
			\textbf{Step 3}: Transform $\boldsymbol{\mu}^{(c)}$ to a new representation $\hat{\boldsymbol{\mu}}^{(c)}$. \;		
			\ForEach {$c$ in $C$}{
				$\hat{\boldsymbol{\mu}}^{(c)}= \overline{\boldsymbol{\mu}} + (c+1) S  \frac{\vec{V}(\overline{\boldsymbol{\mu}}, \boldsymbol{\mu^{(c)}})}{||\overline{\boldsymbol{\mu}}-\boldsymbol{\mu}^{(c)}||}$
			}
			\caption{Initializing $\boldsymbol{\mu}^c$ and $\boldsymbol{\sigma}^c$ for each class.}
			\label{algo:separatedZAlgo}
		\end{algorithm}

 We use Algorithm \ref{algo:separatedZAlgo} from \cite{dinh2022balanced} to initialize the values $\hat{\boldsymbol{\mu}}^c$ and $\boldsymbol{\sigma}^c$. This algorithm aims to disperse the means of different classes in the latent space to more separated positions.  The first step in Algorithm 1 is to use Principle Component Analysis (PCA) \cite{R2} to project the original data into a new representation with the dimension ($d_z$) equal to the dimension of the latent vector ($\textbf{z}^i$). The second step is to calculate the mean ($\boldsymbol{\mu}^c$), the standard deviation ($\boldsymbol{\sigma}^c$) for the Gaussian distribution of each class, and the center of means of classes ($\overline{\boldsymbol{\mu}}$). The third step is to shift the mean  (${\boldsymbol{\mu}}^{(c)}$) calculated in step 2 to a new position ($\hat{\boldsymbol{\mu}}^{(c)}$) so that the mean of each class is more separated from other classes. The reason for this is that the Gaussian distributions of different classes are sometimes mixed. In our implementation, $\boldsymbol{\mu}^{(c)}$ is shifted by  moving along a direction of vector $\vec{V}(\overline{\boldsymbol{\mu}}, \boldsymbol{\mu}^{(c)})$. Fig. \ref{fig:algoScaledLatentSpace} presents an example of shifting $\boldsymbol{\mu}^{(c)}$ to a new position in the latent space. In this figure, the dataset has three classes. In the original dataset, the mean of three classes is closer to the origin.  After using  Algorithm \ref{algo:separatedZAlgo}, four mean samples are moved further from the origin and they are more separated.

The loss function of CTVAE (Eq. (\ref{eq:loss_ctvae})) is formed by adding two new regularizers to the loss function of TVAE. These regularizers encourage the CTVAE to separate the representation of $\textbf{z}$ and $\hat{\textbf{z}}$ for each class. 

\begin{equation}
\label{eq:loss_ctvae}
   \begin{aligned}	
		\ell_{\mbox{CTVAE}}=\frac{1}{n} \sum_{i=1}^{n} \big{[} (\textbf{x}^{(i)}-\hat{\textbf{x}}^{(i)})^{2} +  \beta_1 (\textbf{z}^{(i)}-\hat{\textbf{z}}^{(i)})^{2} \\
		+ \frac{\beta_2}{2} \sum_{j=1}^{J}\big(-1-\log( ({\sigma}^{(i)}_{j})^{2} )+({\mu}^{(i)}_{j})^{2}+({\sigma}^{(i)}_{j})^{2}\big) \\
		+  \beta_3 (\textbf{z}^{(i)}- f(\textbf{z}^{(i)}))^{2} + \beta_4 (\hat{\textbf{z}}^{(i)} - f(\hat{\textbf{z}}^{(i)}))^2 \big{]},
	\end{aligned}
\end{equation}
	where $\textbf{x}^{(i)}$ is the $i^{th}$ data sample in the training set, and $\hat{\textbf{x}}^{(i)}$ is the representation at the output layer of the Hermaphrodite. $\textbf{z}^{(i)}$ is the representation of $\textbf{x}^{(i)}$ in the latent space whilst $\hat{\textbf{z}}^{(i)}$ is the representation of the output layer of the Decoder. ${\mu}^{(i)}_j$ and ${\sigma}^{(i)}_j$ are the $j^{th}$ components of the vector representation of two output layers, i.e., ${\boldsymbol{\mu}^{(i)}}$ and $\boldsymbol{\sigma}^{(i)}$, respectively, of the Encoder of the CTVAE, as observed in Figure~\ref{fig:tvae_fig}. $J$ denotes the dimensionality of the latent space $\textbf{z}^{i}$, and $n$ is the number of samples in the training set. $\beta_1$, $\beta_2$, $\beta_3$, and $\beta_4$ are hyper-parameters.
    $f(\textbf{z}^{(i)})$, $f(\hat{\textbf{z}}^{(i)})$ are functions that match data samples, i.e.,  $\textbf{z}^{(i)}$, $\hat{\textbf{z}}^{(i)}$, to the mean of their class, i.e., ($f(\textbf{z}^{(i)})$ and  $f(\hat{\textbf{z}}^{(i)}) \in \{ \hat{\boldsymbol{\mu}}^{(c)}, c =\{1...|C|\}\} $, $C$ is set of classes).

The first three terms in the loss function of CTVAE are the same as those of TVAE. The fourth term $(\textbf{z}^{(i)}- f(\textbf{z}^{(i)}))^2$ manages to shrink the latent space $\textbf{z}^{(i)}$ into its mean, i.e., $\hat{\boldsymbol{\mu}}^{(c)}$. This prevents the latent space $\textbf{z}^{(i)}$ of a class from being overlapped with data samples of other classes. Similarly, the fifth term $(\hat{\textbf{z}}^{(i)} - f(\hat{\textbf{z}}^{(i)}))^{2}$ also shrinks the reconstruction representation $\hat{\textbf{z}}^{(i)}$ of each class into its corresponding mean, i.e., $\hat{\boldsymbol{\mu}}^{(c)}$. 

To facilitate the classifiers, the previous AE variants \cite{R1, R10, R13, R16, R27, R18, R19} add the regularized terms in the loss function of the AE. Therefore, the data representation of AE is influenced by both reconstruction and regularized terms. The trade-off between the reconstruction and regularized terms during the training process of the AE may lower the classification accuracy. 
Unlike the regularized terms used in the AE variants, the CTVAE transfers the input data  $\textbf{x}^{(i)}$ into the latent space $\boldsymbol{\mu}^{(i)}$, and then the data samples $\textbf{z}^{(i)}$ are generated from separated Gaussian distributions. Note that, the data samples of a class may be close to its mean, whilst the data samples of different classes are separated from each other. Next, the Hermaphrodite of the CTVAE maps the data $\textbf{z}^{(i)}$ into $\hat{\textbf{x}}^{(i)}$ before the Decoder maps the data $\hat{\textbf{x}}^{(i)}$ to $\hat{\textbf{z}}^{(i)}$. The training process of the CTVAE involves two processes, e.g., making data $\hat{\textbf{x}}^{(i)}$  as close to ${\textbf{x}}^{(i)}$ as possible and making data $\hat{\textbf{z}}^{(i)}$  as close to ${\textbf{z}}^{(i)}$ as possible. After the training process, we feed training data and testing data into the Decoder to obtain the data $\hat{\textbf{z}}^{(i)}$ as the data representation of the ${\textbf{x}}^{(i)}$. Note that, $\hat{\textbf{z}}^{(i)}$ is referred to as the \textit{reconstruction representation}, and it is used as the input of classifiers in the IDSs~\footnote{One can question why $\textbf{z}^{(i)}$ of the CTVAE is not used as an input for the classifiers. The CTVAE uses the class label, i.e., $c$, to generate $\textbf{z}^{(i)}$ from its mean, i.e., $\hat{\boldsymbol{\mu}}^{(c)}$. For the inference, there is no label information.}.

\section{EXPERIMENTAL SETTINGS}
\label{experiment_setting}
\subsection{Performance Metrics}
We compare the proposed model with popular machine learning algorithms and the state-of-the-art methods including MAE~\cite{R1}, MVAE~\cite{R1}, CSAEC~\cite{R18}, and XgBoost~\cite{chen2016xgb} using two popular metrics: Accuracy and F-Score. 	The accuracy metric is calculated as follows:

\begin{equation}
{Accuracy} = \frac{\emph{TP}+\emph{TN}}{\emph{TP}+\emph{TN}+\emph{FP}+\emph{FN}},
\label{eq:bloc_first}
\end{equation}

\noindent where $\emph{TP}$ and $\emph{FP}$ are the number of correct and incorrect predicted samples for the attack class, respectively, and $\emph{TN}$ and $\emph{FN}$ are the number of corrected and incorrect predicted samples for the normal classes.

The F-score is the harmonic mean of the precision and the recall, which is an effective measurement for imbalanced datasets:
\begin{equation}
{F\emph{-}score} = 2 \times \frac{{Precision} \times {Recall} }{{Precision} +{Recall}} ,
\label{eq:bloc_last}
\end{equation}

\noindent where Precision is the ratio of the number of correct detections to the number of incorrect detections:
\begin{equation}
{Precision} = \frac{\emph{TP}}{\emph{TP}+\emph{FP}} ,
\end{equation}
\noindent and Recall is the ratio of the number of correct detections to the number of false detections:
\begin{equation}
{Recall} = \frac{\emph{TP}}{\emph{TP}+\emph{FN}} \cdot
\end{equation}

To further reinforce the theoretical results/contributions (in order to evaluate the quality of the data representation of the CTVAE), we use the two metrics \cite{R22}, i.e., the distance between means of different classes (between-class variance, i.e., $d_{bet}$) and distance between data samples of a class to its mean  (within-class variance, i.e., $d_{wit}$). If the value of $d_{bet}$ is large, the means of classes are far from each other (better separation and easier for classifiers).  If the value of $d_{wit}$ is small, it means that the data samples of a class are forced to its mean.
		
		The average between-class variance is calculated as follows:
		\begin{equation}
		\label{eq:d_between_class_Big}
		\begin{aligned}	
		\textbf{B} = \frac{1}{2}\sum_{c=1}^{|C|} \sum_{c^{'}=1}^{|C|} |\boldsymbol{\mu}^{(c)} - \boldsymbol{\mu}^{(c^{'})}|^2 ,
		\end{aligned}
		\end{equation} 
		\begin{equation}
		\label{eq:d_between_class_small}
		\begin{aligned}	
		d_{bet} = \frac{1}{d_\textbf{B}}\sum_{k=1}^{d_\textbf{B}} {B}_k,
		\end{aligned}
		\end{equation}
		where $C$ is a set of classes,  $|C|$ is the number of classes. $c, c^{'} \in C$ are class labels. $\boldsymbol{\mu}^{(c)}$, $\boldsymbol{\mu}^{(c^{'})}$ are the means of classes, i.e., $c$ and $c^{'}$, respectively. $d_\textbf{B}$ is the dimensionality of $\textbf{B}$. ${B}_k$ is the $k^{th}$ component of vector $\textbf{B}$.
		
		The average within-class variance is calculated as follows:
		\begin{equation}
		\label{eq:d_within_class_big}
		\begin{aligned}	
		\textbf{T} = \frac{1}{n}\sum_{c=1}^{|C|} \sum_{j=1}^{n_c} |\textbf{x}^{(j,c)} - \boldsymbol{\mu}^{(c)}|^2 ,
		\end{aligned}
		\end{equation}
		
		\begin{equation}
		\label{eq:d_within_class_small}
		\begin{aligned}	
		d_{wit} = \frac{1}{d_\textbf{T}}\sum_{k=1}^{d_\textbf{T}} {T}_k,
		\end{aligned}
		\end{equation}
		 where $\textbf{x}^{(j,c)}$ is the $j^{th}$ data sample of class $c$, and $n_c$ is the number of samples of class $c$. $\boldsymbol{\mu}^{(c)}$ is the means of class $c$. $n$ is the size of the dataset. $d_{\textbf{T}}$ is the dimensionality of $\textbf{T}$. ${T}_k$ is the $k^{th}$ component of vector $\textbf{T}$.  
		
		Our design of CTVAE aims to make $d_{bet}$ as large as possible and $d_{wit}$ as small as possible. This is because Algorithm 1 in the revised paper tries to increase the value of $d_{bet}$ by moving the means of classes away from each other. In addition, the loss function of the CTVAE tries to decrease the value of $d_{wit}$ by forcing the data samples in the representation space into the mean of each class.

\subsection{Datasets}
\begin{table}[htp]
	\centering
	\setlength\tabcolsep{3.5pt}
	\caption{Datasets information.}
	\label{tab:tab_iot_datasets}
	\begin{tabular}{|c|c|c|c|c|c|c|}
		\hline
		\textbf{\begin{tabular}[c]{@{}c@{}}Attack\\ Types\end{tabular}} & \textbf{Labels}& \textbf{\begin{tabular}[c]{@{}c@{}}Danmini\\ Doorbell\end{tabular}} & \textbf{Ecobee} & \textbf{Philips} & \textbf{\begin{tabular}[c]{@{}c@{}}737E\\ Security\\ Camera\end{tabular}} & \textbf{\begin{tabular}[c]{@{}c@{}}838\\ Security\\ Camera\end{tabular}} \\ \hline
		& \begin{tabular}[c]{@{}c@{}}Benign\\ Traffic\end{tabular} & 49548& 13113& 175240& 62154& 98514\\ \hline
		& Scan& 29849& 27494& 27859 & 29297& 28397\\ \cline{2-7} 
		& UDP & 105874& 104791 & 105782& 104011& 104658 \\ \cline{2-7} 
		& Combo& 59718& 53012& 58152 & 61380& 57530\\ \cline{2-7} 
		& Junk& 29068& 30312& 28349 & 30898& 29068\\ \cline{2-7} 
		\multirow{-5}{*}{Gafgyt}& TCP & 92141& 95021& 92581 & 104510& 89387\\ \hline
		& ACK & 102195& 113285 & 91123 & 60554& 57997\\ \cline{2-7} 
		& Scan& 107685& 43192& 103621& 96781& 97096\\ \cline{2-7} 
		& \begin{tabular}[c]{@{}c@{}}UDP\\ Plain\end{tabular}& 81982& 87368& 80808 & 56681& 53785\\ \cline{2-7} 
		& SYN & 122573& 116807 & 118128& 65746& 61851\\ \cline{2-7} 
		\multirow{-5}{*}{ Mirai}& UDP & 237665& 151481 & 217034& 156248& 158608 \\ \hline
	\end{tabular}
\end{table}
\begin{table}[htp]
	\centering
	\setlength\tabcolsep{1.5pt}
	\caption{The ID of datasets.}
	\label{tab:tab_iot_id}
	\begin{tabular}{|c|c|c|}
\hline
\textbf{Names}& \textbf{No. classes} & \textbf{ID} \\ \hline
(DanG-6) Gafgyt trained IoT-01 & 6& IoT-01 \\ \hline
(PhiG-6) Gafgyt trained IoT-02 & 6& IoT-02 \\ \hline
(838G-6) Gafgyt trained IoT-03 & 6& IoT-03 \\ \hline
(EcoG-6) Gafgyt trained IoT-04 & 6& IoT-04 \\ \hline
(737G-6) Gafgyt trained IoT-05 & 6& IoT-05 \\ \hline
(EcoMG-11) Mirai and Gafgyt trained IoT-06   & 11& IoT-06 \\ \hline
(838MG-11) Mirai and Gafgyt trained IoT-07   & 11& IoT-07 \\ \hline
(737GUC-2) Gafgyt (UDP,Combo) trained IoT-08 & 2& IoT-08 \\ \hline
(838GUC-2) Gafgyt (UDP,Combo) trained IoT-09 & 2& IoT-09 \\ \hline
(DanM-2) Mirai trained IoT-10& 2& IoT-10 \\ \hline
(838M-2) Mirai trained IoT-11& 2& IoT-11 \\ \hline
\end{tabular}
	
\end{table}

	\begin{table}[t]
		\centering
		\setlength\tabcolsep{1.5pt}
		\caption{{IDS binary datasets.}}
		\label{tab:tab_ids_datasets4}
		\begin{tabular}{|c|cc|cc|}
			\hline
			\multirow{2}{*}{\textbf{Datasets}} & \multicolumn{2}{c|}{\textbf{No. Train}}& \multicolumn{2}{c|}{\textbf{No. Test}} \\ \cline{2-5} 
			& \multicolumn{1}{c|}{\textbf{benign}} & \textbf{attack} & \multicolumn{1}{c|}{\textbf{benign}} & \textbf{attack} \\ \hline
			UNSW-NB15    & \multicolumn{1}{c|}{37000}& 45332& \multicolumn{1}{c|}{56000}& 119341 \\ \hline
			IDS2017 & \multicolumn{1}{c|}{219068} & 46859& \multicolumn{1}{c|}{153091} & 32806\\ \hline
		\end{tabular}
	\end{table}


In order to perform the evaluations, we use various types of datasets. First, from the original dataset in Table \ref{tab:tab_iot_datasets}, we form 11 datasets \cite{R13} to evaluate the proposed model in terms of high dimensions and redundant datasets. These 11 datasets are presented in Table II. They are divided into binary datasets and multi-class datasets. In binary datasets, we randomly select 70\% and 30\% of normal traffic for training and testing, respectively. For the attack data, we create two sub-datasets as follows. First, for IoT-10 and IoT-11, the Mirai botnet is used for training, respectively, and the Gafgyt botnet is used for testing. Second, for IoT-08 and IoT-09 datasets, the Combo and UDP attacks in the Gafgyt are used for training, and the remaining of the Gafgyt and the Mirai botnet are used for testing. It is worth noting that in these datasets,	there is always a new type of attack in the testing sets. For multi-class classification, we create two scenarios. The first scenario, from IoT-01 to IoT-05 datasets, includes the normal traffic and five	types of Gafgyt attacks. The second scenario, IoT-06 and IoT-07 datasets, consists of normal traffic, five types of Mirai attacks, and five types of Gafgyt attacks. For each scenario, 70\% samples are for training and 30\% are for testing.
		
		Second, we use two binary datasets, e.g., IDS-2017 \cite{IDS2017} and UNSW-NB15 \cite{UNSW} as shown in Table \ref{tab:tab_ids_datasets4} to evaluate the proposed model in dealing with imbalanced data. Both datasets have two types of data, e.g., normal and attack. We assume that the normal data samples in the training set have multiple distributions. We try to label the normal data samples in the training set by using Kmean and silhouette values to find the reasonable distributions of the normal samples \cite{R29}. After that, the CTVAE can train by a new multi-label training set which is less imbalanced between the normal and the attack samples.

\subsection{Parameter Settings}
The experiments are implemented in Python using two frameworks Tensorflow and Scikit-learn~\cite{Scikit}.  The ADAM optimization algorithm \cite{kingma2017adam} is used to train the neural networks. The learning rate $\alpha$ is initially set at \textit{$10^{-4}$}. The weights are initialized using  Glorot's method \cite{glorot10a} to facilitate the convergence. The number $epochs$ is set at 2000 and the  $batch\_size$ is set at 100.  

The number of neurons in the hidden layers of all neural network models, i.e., MVAE, MAE, and CTVAE is shown in Table \ref{tab:neural_architecture}. For example, the dimensionality of latent space ($d_z$) is $10$ which is calculated using the rule ($d_z \approx \sqrt{d_{input}}$) in \cite{R1, R12}, where $d_{input}$ is dimensionality of the input. The dimensionality of the hidden layers, i.e., $h1$, $h2$, and $h3$, is approximated by half of $d_{input}$. In addition, the dimensionality of $\boldsymbol{\mu}$ and ${\boldsymbol{\sigma}}$ of two models, i.e., MVAE and CTVAE, are the same as that of $\textbf{z}$. Note that, in order to compare the performance of three models, i.e., MAE, MVAE, and CTVAE, we attempt to configure the same parameter settings. The additional settings of the CTVAE are two layers, e.g., $h3$ and $\hat{\textbf{z}}$.
For MAE and MVAE, the value of $\boldsymbol{\mu}_{\textbf{y}^{(i)}}$ is set at {2} for a normal class and {3} for an abnormal class in binary datasets as in~\cite{R1}. The value of $\boldsymbol{\mu}_{\textbf{y}^{(i)}}$ is set from {2} to {$|C| + 2$} in the multiclass datasets, where $|C|$ is the number of classes. After training the RL models, the Random Forest (RF) is applied to their  representation to make the final prediction~\footnote{We also tested the combination of RL methods with other machine learning models including SVM, RL, and DT. The results are similar to RF and they are not presented for the sake of the concise presentation.}.

\begin{table}[ht]
	\centering
	\caption{Number of neurons in the MAE, MVAE, and CTVAE.}
	\label{tab:neural_architecture}
	\begin{adjustbox}{max width=\textwidth}
		\begin{tabular}{|c|c|c|c|c|c|c|c|c|}
			\hline
			Methods&Input&\textit{h1}&$ \boldsymbol{\mu}, \boldsymbol{\sigma}, \textbf{z} $&\textit{h2}&$\hat{\textbf{x}}$&\textit{h3}&$\hat{\textbf{z}}$ \\ \hline
			MAE&115&50&10&50&115&-&- \\ \hline
			MVAE&115&50&10&50&115&-&- \\ \hline
			CTVAE&115&50&10&50&115&50&10 \\ \hline
		\end{tabular}
	\end{adjustbox}
\end{table}
\begin{table}[ht]
	\centering
	\caption{Parameters for the grid search.}
	\label{tab:grid_configurations}
	\begin{adjustbox}{max width=\textwidth}
		\begin{tabular}{|l|l|}
			\hline
			LR  & \begin{tabular}[c]{@{}l@{}}Default or $solver$=\{'lbfgs', 'liblinear'\};\\  $C$=\{0.1, 0.5, 1.0, 5.0, 10.0\}\end{tabular}\\ \hline
			SVM & LinearSVC: default or $C$=\{0.1, 0.2, 0.5, 1.0, 5.0, 10.0\} \\ \hline
			DT  & \begin{tabular}[c]{@{}l@{}}Default or  $criterion$=\{'gini', 'entropy'\};\\  $max\_depth$=\{5, 10, 20, 50, 100\}\end{tabular} \\ \hline
			RF  & Default or  $n\_estimators$=\{5, 10, 20, 50, 100, 150\} \\ \hline
			Xgb & \begin{tabular}[c]{@{}l@{}}Default or $min\_child\_weight$=\{5, 10\},  \\ $gamma$=\{0.5, 1.5\}, $subsample$=\{1.0, 0.9\}\end{tabular} \\ \hline
		\end{tabular}
	\end{adjustbox}
\end{table}
For CSAE models, the same architecture in \cite{park2020low} is built using 1D convolution. The CSAE first is trained in an unsupervised manner. After that, the encoder of CSAE is used as a pre-trained model to train the 1D Convolutional Neural Network (1D-CNN). After training the 1D-CNN model in a supervised learning mode, we use the data representation of the 1D-CNN at the layer which is located right in front of the softmax layer of the 1D-CNN \cite{phan2018Tree}. The combination of CSAE and 1D-CNN is as referred to CSAEC. For the Xgboost (Xgb) classifier, we conducted a grid search to tune its hyper-parameters. Similarly, the grid search is also used for the four classifiers, i.e., LR, SVM \cite{gridSVM2016}, DT \cite{gridDT2019}, and RF \cite{gridRF2019} are shown in Table \ref{tab:grid_configurations}.
Finally, for the proposed models, we set {1} for all hyper-parameters in the loss functions, i.e., $\beta_1$, $\beta_2$, $\beta_3$ and $\beta_4$. 

\section{RESULTS AND ANALYSIS}
\label{result_analysis}
In this section, we will evaluate the performances of CTVAE on the binary and multi-class datasets. The results are compared with the state-of-the-art methods in the same fields. We also investigate various characteristics of CTVAE including its model complexity and the properties of the latent representation  and the reconstruction representation.

\subsection{Performance on Multi-class Datasets}
\label{l_ctvae_result}
\begin{table*}[t]
	\centering
	\setlength\tabcolsep{10pt}
	\small
	\caption{Results on IoT multi-class datasets.}
	\label{tab:tab_multi_label_TVAE_vs_ALL}
	\begin{tabular}{|c|c|c|ccccccc|}
		\hline 
		& && \multicolumn{7}{c|}{\textbf{Multiple classes datasets}}   \\ \cline{4-10} 
		\multirow{-2}{*}{} & \multirow{-2}{*}{\textbf{Clf}} & \multirow{-2}{*}{\textbf{Methods}} & \multicolumn{1}{c|}{\textbf{IoT-01}}& \multicolumn{1}{c|}{\textbf{IoT-02}}& \multicolumn{1}{c|}{\textbf{IoT-03}}& \multicolumn{1}{c|}{\textbf{IoT-04}}& \multicolumn{1}{c|}{\textbf{IoT-05}}& \multicolumn{1}{c|}{\textbf{IoT-06}}& \textbf{IoT-07}\\ \hline \hline
		& Xgb   & STA & \multicolumn{1}{c|}{92.3}& \multicolumn{1}{c|}{94.3}& \multicolumn{1}{c|}{93.2}& \multicolumn{1}{c|}{91.0}& \multicolumn{1}{c|}{92.1}& \multicolumn{1}{c|}{94.1}& 95.9\\ \cline{2-10} 
		& LR& STA & \multicolumn{1}{c|}{67.0}& \multicolumn{1}{c|}{75.1}& \multicolumn{1}{c|}{70.8}& \multicolumn{1}{c|}{61.2}& \multicolumn{1}{c|}{65.4}& \multicolumn{1}{c|}{84.0}& 85.4\\ \cline{2-10} 
		& SVM   & STA & \multicolumn{1}{c|}{67.0}& \multicolumn{1}{c|}{75.1}& \multicolumn{1}{c|}{70.8}& \multicolumn{1}{c|}{63.0}& \multicolumn{1}{c|}{65.5}& \multicolumn{1}{c|}{83.5}& 85.6\\ \cline{2-10} 
		& DT& STA & \multicolumn{1}{c|}{67.2}& \multicolumn{1}{c|}{70.0}& \multicolumn{1}{c|}{74.0}& \multicolumn{1}{c|}{65.4}& \multicolumn{1}{c|}{65.6}& \multicolumn{1}{c|}{81.4}& 86.7\\ \cline{2-10} 
		& RF& STA & \multicolumn{1}{c|}{67.2}& \multicolumn{1}{c|}{75.2}& \multicolumn{1}{c|}{70.9}& \multicolumn{1}{c|}{61.3}& \multicolumn{1}{c|}{65.6}& \multicolumn{1}{c|}{85.0}& 85.8\\ \cline{2-10} 
		& & CSAEC & \multicolumn{1}{c|}{92.0}& \multicolumn{1}{c|}{94.2}& \multicolumn{1}{c|}{92.8}& \multicolumn{1}{c|}{90.6}& \multicolumn{1}{c|}{92.1}& \multicolumn{1}{c|}{92.3}& 96.5\\ \cline{3-10} 
		& & MAE& \multicolumn{1}{c|}{92.1}& \multicolumn{1}{c|}{94.1}& \multicolumn{1}{c|}{93.4}& \multicolumn{1}{c|}{90.9}& \multicolumn{1}{c|}{92.2}& \multicolumn{1}{c|}{78.1}& 94.2\\ \cline{3-10} 
		& & MVAE & \multicolumn{1}{c|}{67.2}& \multicolumn{1}{c|}{74.2}& \multicolumn{1}{c|}{70.5}& \multicolumn{1}{c|}{60.0}& \multicolumn{1}{c|}{65.7}& \multicolumn{1}{c|}{62.8}& 76.8\\ \cline{3-10} 
		\multirow{-9}{*}{Acc} & \multirow{-4}{*}{RF}& \cellcolor[HTML]{DDDDDD}CTVAE & \multicolumn{1}{c|}{\cellcolor[HTML]{DDDDDD}\textbf{93.0}} & \multicolumn{1}{c|}{\cellcolor[HTML]{DDDDDD}\textbf{95.0}} & \multicolumn{1}{c|}{\cellcolor[HTML]{DDDDDD}\textbf{93.9}} & \multicolumn{1}{c|}{\cellcolor[HTML]{DDDDDD}\textbf{91.8}} & \multicolumn{1}{c|}{\cellcolor[HTML]{DDDDDD}\textbf{93.1}} & \multicolumn{1}{c|}{\cellcolor[HTML]{DDDDDD}\textbf{94.5}} & \cellcolor[HTML]{DDDDDD}\textbf{96.6} \\ \hline \hline
		& Xgb   & STA & \multicolumn{1}{c|}{89.4}& \multicolumn{1}{c|}{92.2}& \multicolumn{1}{c|}{90.9}& \multicolumn{1}{c|}{87.9}& \multicolumn{1}{c|}{89.1}& \multicolumn{1}{c|}{92.9}& 94.5\\ \cline{2-10} 
		& LR& STA & \multicolumn{1}{c|}{55.3}& \multicolumn{1}{c|}{66.2}& \multicolumn{1}{c|}{60.3}& \multicolumn{1}{c|}{47.5}& \multicolumn{1}{c|}{53.4}& \multicolumn{1}{c|}{78.7}& 80.3\\ \cline{2-10} 
		& SVM   & STA & \multicolumn{1}{c|}{55.3}& \multicolumn{1}{c|}{66.2}& \multicolumn{1}{c|}{60.3}& \multicolumn{1}{c|}{52.9}& \multicolumn{1}{c|}{53.5}& \multicolumn{1}{c|}{78.3}& 80.5\\ \cline{2-10} 
		& DT& STA & \multicolumn{1}{c|}{55.7}& \multicolumn{1}{c|}{63.5}& \multicolumn{1}{c|}{65.8}& \multicolumn{1}{c|}{54.7}& \multicolumn{1}{c|}{53.6}& \multicolumn{1}{c|}{77.4}& 82.4\\ \cline{2-10} 
		& RF& STA & \multicolumn{1}{c|}{55.7}& \multicolumn{1}{c|}{66.4}& \multicolumn{1}{c|}{60.4}& \multicolumn{1}{c|}{47.7}& \multicolumn{1}{c|}{53.6}& \multicolumn{1}{c|}{79.7}& 80.7\\ \cline{2-10} 
		& & CSAEC & \multicolumn{1}{c|}{88.9}& \multicolumn{1}{c|}{91.8}& \multicolumn{1}{c|}{89.9}& \multicolumn{1}{c|}{87.0}& \multicolumn{1}{c|}{89.0}& \multicolumn{1}{c|}{91.8}& 95.1\\ \cline{3-10} 
		& & MAE& \multicolumn{1}{c|}{89.2}& \multicolumn{1}{c|}{91.9}& \multicolumn{1}{c|}{91.3}& \multicolumn{1}{c|}{87.6}& \multicolumn{1}{c|}{89.3}& \multicolumn{1}{c|}{76.9}& 94.2\\ \cline{3-10} 
		& & MVAE & \multicolumn{1}{c|}{64.9}& \multicolumn{1}{c|}{71.5}& \multicolumn{1}{c|}{67.9}& \multicolumn{1}{c|}{57.0}& \multicolumn{1}{c|}{62.8}& \multicolumn{1}{c|}{61.7}& 75.5\\ \cline{3-10} 
		\multirow{-9}{*}{Fscore} & \multirow{-4}{*}{RF}& \cellcolor[HTML]{DDDDDD}CTVAE & \multicolumn{1}{c|}{\cellcolor[HTML]{DDDDDD}\textbf{90.9}} & \multicolumn{1}{c|}{\cellcolor[HTML]{DDDDDD}\textbf{93.5}} & \multicolumn{1}{c|}{\cellcolor[HTML]{DDDDDD}\textbf{92.2}} & \multicolumn{1}{c|}{\cellcolor[HTML]{DDDDDD}\textbf{89.5}} & \multicolumn{1}{c|}{\cellcolor[HTML]{DDDDDD}\textbf{91.1}} & \multicolumn{1}{c|}{\cellcolor[HTML]{DDDDDD}\textbf{93.3}} & \cellcolor[HTML]{DDDDDD}\textbf{95.5} \\ \hline
	\end{tabular}
	
\end{table*}

\begin{table}[t]
	\centering
	\setlength\tabcolsep{2.0pt}
	\small
	\caption{Results on binary datasets.}
	\label{tab:tab_result_anomaly_detection}
	\begin{tabular}{|c|c|c|cccc|}
		\hline
		& && \multicolumn{4}{c|}{\textbf{Anomaly datasets}}    \\ \cline{4-7} 
		\multirow{-2}{*}{\textbf{}} & \multirow{-2}{*}{\textbf{Clf}} & \multirow{-2}{*}{\textbf{Methods}} & \multicolumn{1}{c|}{\textbf{IoT-08}} & \multicolumn{1}{c|}{\textbf{IoT-09}}& \multicolumn{1}{c|}{\textbf{IoT-10}}& \textbf{IoT-11}\\ \hline \hline
		& Xgb   & STA & \multicolumn{1}{c|}{82.7} & \multicolumn{1}{c|}{69.8}& \multicolumn{1}{c|}{99.6}& 41.7\\ \cline{2-7} 
		& LR& STA & \multicolumn{1}{c|}{99.0} & \multicolumn{1}{c|}{98.3}& \multicolumn{1}{c|}{39.7}& 42.0\\ \cline{2-7} 
		& SVM   & STA & \multicolumn{1}{c|}{97.5} & \multicolumn{1}{c|}{97.1}& \multicolumn{1}{c|}{40.1}& 41.6\\ \cline{2-7} 
		& DT& STA & \multicolumn{1}{c|}{79.3} & \multicolumn{1}{c|}{74.8}& \multicolumn{1}{c|}{38.1}& 42.7\\ \cline{2-7} 
		& RF& STA & \multicolumn{1}{c|}{97.6} & \multicolumn{1}{c|}{95.3}& \multicolumn{1}{c|}{40.0}& 42.1\\ \cline{2-7} 
		& & CSAEC & \multicolumn{1}{c|}{90.0} & \multicolumn{1}{c|}{80.2}& \multicolumn{1}{c|}{89.9}& 87.9\\ \cline{3-7} 
		& & MAE & \multicolumn{1}{c|}{95.5} & \multicolumn{1}{c|}{94.8}& \multicolumn{1}{c|}{98.8}& 42.1\\ \cline{3-7} 
		& & MVAE& \multicolumn{1}{c|}{88.5} & \multicolumn{1}{c|}{90.1}& \multicolumn{1}{c|}{95.5}& 61.0\\ \cline{3-7} 
		\multirow{-9}{*}{Acc} & \multirow{-4}{*}{RF}& \cellcolor[HTML]{DDDDDD}CTVAE & \multicolumn{1}{c|}{\cellcolor[HTML]{DDDDDD}\textbf{100.0}} & \multicolumn{1}{c|}{\cellcolor[HTML]{DDDDDD}\textbf{98.9}} & \multicolumn{1}{c|}{\cellcolor[HTML]{DDDDDD}\textbf{99.7}} & \cellcolor[HTML]{DDDDDD}\textbf{99.7} \\ \hline \hline
		& Xgb   & STA & \multicolumn{1}{c|}{88.3} & \multicolumn{1}{c|}{78.3}& \multicolumn{1}{c|}{99.6}& 50.4\\ \cline{2-7} 
		& LR& STA & \multicolumn{1}{c|}{99.1} & \multicolumn{1}{c|}{98.4}& \multicolumn{1}{c|}{52.0}& 50.8\\ \cline{2-7} 
		& SVM   & STA & \multicolumn{1}{c|}{97.8} & \multicolumn{1}{c|}{97.4}& \multicolumn{1}{c|}{52.5}& 50.3\\ \cline{2-7} 
		& DT& STA & \multicolumn{1}{c|}{86.1} & \multicolumn{1}{c|}{82.0}& \multicolumn{1}{c|}{50.3}& 51.5\\ \cline{2-7} 
		& RF& STA & \multicolumn{1}{c|}{97.9} & \multicolumn{1}{c|}{96.0}& \multicolumn{1}{c|}{52.4}& 50.9\\ \cline{2-7} 
		& & CSAEC & \multicolumn{1}{c|}{92.8} & \multicolumn{1}{c|}{85.7}& \multicolumn{1}{c|}{92.3}& 89.7\\ \cline{3-7} 
		& & MAE & \multicolumn{1}{c|}{96.4} & \multicolumn{1}{c|}{95.7}& \multicolumn{1}{c|}{98.9}& 50.9\\ \cline{3-7} 
		& & MVAE& \multicolumn{1}{c|}{91.9} & \multicolumn{1}{c|}{92.2}& \multicolumn{1}{c|}{96.0}& 69.4\\ \cline{3-7} 
		\multirow{-9}{*}{Fscore} & \multirow{-4}{*}{RF}& \cellcolor[HTML]{DDDDDD}CTVAE & \multicolumn{1}{c|}{\cellcolor[HTML]{DDDDDD}\textbf{100.0}} & \multicolumn{1}{c|}{\cellcolor[HTML]{DDDDDD}\textbf{99.0}} & \multicolumn{1}{c|}{\cellcolor[HTML]{DDDDDD}\textbf{99.8}} & \cellcolor[HTML]{DDDDDD}\textbf{99.7} \\ \hline
	\end{tabular}
	
\end{table}

This section compares the performance of CTVAE with the machine learning models and the RL methods on multi-class datasets. The tested machine learning methods include  Xgb, LR, SVM, DT, and RF, which are trained on the original data. These methods are considered as the stand-alone classifiers (STA).  The tested RL methods includes CSAEC~\cite{R18}, MAE~\cite{R1}, and MVAE~\cite{R1}. 

Table~\ref{tab:tab_multi_label_TVAE_vs_ALL} presents the performance of CTVAE compared to the other methods on multi-class datasets. 
It can be seen from the table that CTVAE is often much better than those of the four STAs, i.e., RF, DT, SVM, and LR,  on both metrics: $Accuracy$ and $F\mbox{-}score$. For example, on IoT-01, CTVAE achieves accuracy of 93.0\%, while this value of LR, SVM, DT, and RF is only around 67\%. For the Fscore metric, the value of CTVAE is 90.9\% while the values of LR, SVM, DT, and RF are only about 66\%. The results on the other datasets are also similar to the results on the IoT-01. Among the five STAs, the results of Xgb are much better than the rest. However, the results of  Xgb are still lower than those of the CTVAE. Moreover, the training and testing time of Xgb is also much longer than CTVAE (see Subsection~\ref{l_complexity}). Therefore, this model is not appropriate for deploying on the IoT gateways. 

Comparing between representation methods in Table~\ref{tab:tab_multi_label_TVAE_vs_ALL}. As can be seen, CTVAE achieves the best result on all the datasets. For example on the IoT-02 dataset, the accuracy of CTVAE is 95\%, while the accuracy of CSAEC, MAE, and MVAE   are only 94.2\%, 94.1\%, and 74.2\%, respectively. Similarly, the F-score of CSAEC is always higher than those of the others. On the IoT-02 dataset, the F-score of CTVAE is 93.5\% compared to 91.8\%, 91.9\%, and 71.5\% of CSAEC, MAE, and MVAE, respectively. Amongst the four tested methods, the table shows that the performance of MVAE is much worse compared to the rest. The reason is that the latent representation of MVAE is nondeterministic due to the random sampling process to generate the latent representation. Thus, the latent representation of MVAE is not as stable as those of the others.

Overall, the results in this subsection show the superior performance of our proposed model, i.e., CTVAE. The model is trained in a supervised manner and its performance is often higher than those of the state-of-the-art supervised RL models, i.e., MAE and CSAEC. The CTVAE also achieves better results than the well-known ensemble learning method, i.e., Xgboost and popular classifiers that are trained on the original dataset. 

\subsection{Performance on Binary Datasets}

This section compares the performance of CTVAE with the machine learning models and the RL methods on binary datasets. The results are to compare the effectiveness of methods in detecting anomalies.

The results of CTVAE and the other tested methods on binary datasets are presented in Table~\ref{tab:tab_result_anomaly_detection}.  There are some interesting results that can be observed from Table~\ref{tab:tab_result_anomaly_detection}. First, it seems that three datasets (IoT-08, IoT-09, IoT-10) are relatively easy to classify and all tested models achieve good performance. Second, the result of Xgb is worse compared to the other machine learning methods. For example, the accuracy of Xgb on IoT-08 is only 82.7\% while these values of LR, SVM, DT, and RF are 99.0\%, 97.5\%, 79.3\%, and 97.6\%, respectively. The reason is that Xgb is overfitted on these datasets due to its high complexity. Third, the dataset IoT-11 is the most difficult one, and most of the methods perform unconvincingly on this dataset. For example, the accuracy of Xgb, MVAE, and MAE are 41.7\%, 61.0\%, and 42.1 \%, respectively. It means that ML models face difficulties in distinguishing the new Gafgyt attacks from the trained Mirai attacks. Conversely, CTVAE still achieves very high accuracy on IoT-11. Last, CTVAE achieves the best result among all the tested methods. Specifically, the accuracy and Fscore of CVTAE on IoT-10 are 99.7\% and 99.8\%, respectively and these values on IoT-11 are 99.7\%. 
As can be seen, CTVAE, MVAE, and MAE can achieve much better results in comparison with the remaining methods. This is because three methods can separate the benign samples and known attack samples in the training sets. Therefore, the abnormal samples can be differentiated from the benign samples in the testing set. However, the accuracy and Fscore of the CTVAE are greater than those of the start-of-the-art model, i.e., MAE. This is because the MAE is affected by a trade-off between the reconstruction term and the regularized terms during the process of training. Unlike the AE, CTVAE can generate the separated data samples between the benign and the known attacks during the training process. Overall, the results in this section show the better performance of CTVAE compared with the well-known traditional machine learning models and the state-of-the-art RL methods.  

\subsection{Perfomance on Imbalanced Datasets.}

	\begin{table}[t]
		\centering
		\setlength\tabcolsep{1.5pt}
		\caption{{Performance of CTVAE on the imbalanced datasets.}}
		\label{tab:ids-multi-distri-results13}
		\begin{tabular}{|c|cc|cc|}
			\hline
			\multirow{2}{*}{\textbf{Datasets}} & \multicolumn{2}{c|}{\textbf{Accuracy}}   & \multicolumn{2}{c|}{\textbf{Fscore}} \\ \cline{2-5} 
			& \multicolumn{1}{c|}{\textbf{CTVAE-O}} & \textbf{CTVAE-K} & \multicolumn{1}{c|}{\textbf{CTVAE-O}} & \textbf{CTVAE-K} \\ \hline
			UNSW-NB15    & \multicolumn{1}{c|}{0.884} & \textbf{0.893}   & \multicolumn{1}{c|}{0.887} & \textbf{0.909}   \\ \hline
			IDS2017 & \multicolumn{1}{c|}{0.979} & \textbf{0.981}   & \multicolumn{1}{c|}{0.979} & \textbf{0.982}   \\ \hline
		\end{tabular}
		
	\end{table}
	
	To show the effectiveness of the CTVAE for the imbalanced data in IoT IDS systems, we assume that the number of normal samples is significantly greater than those of other attacks. Therefore, the normal data samples may contain numerous distributions. To address this problem, we use KMean and silhouette values [R29] to find the reasonable distributions of the benign data samples. As a result, the CTVAE is trained on the new multi-class datasets that are less imbalanced. The specific configuration of the experiment is as follows. The number of clusters used in the K-Mean algorithm is picked up from a list of $\{2,3,4,5,6,7\}$. The greatest silhouette value shows the numbers of the most reasonable distributions of the benign data samples. The experimental result shows that the number of distributions of the benign samples applied for UNSW-NB15 is $4$, whilst the figure for IDS-2017 is $6$. Therefore, the number of classes in the training set of UNSW-NB15 and IDS2017  is $|C| = 5$ and $|C|=7$, respectively. Note that, the number of classes in the original binary datasets, i.e., UNSW-NB15 and IDS2017, is $|C| = 2$. For example, the IDS2017 dataset has two classes, i.e., ``normal" ($c=0$) and ``attack" ($c=1$), with the numbers of data samples, i.e., 219086 and 46859, respectively. We obtain a new dataset with seven classes ($|C|=7$, $C=\{0,1,2,3,4,5,6\}$), i.e., ``normal0" ($c=0$), ``normal1" ($c=1$), ``normal2" ($c=2$), ``normal3" ($c=3$), ``normal4" ($c=4$), ``normal5" ($c=5$), and ``attack" ($c=6$) with the numbers of data samples, i.e., 51000, 49286, 45171, 29319, 24843, 19449, and 46859, respectively.

	Table \ref{tab:ids-multi-distri-results13} shows the results of CTVAE with two cases on two binary datasets, i.e., UNSW-NB15 and IDS2017. The first case is that CTVAE uses the original binary datasets (CTVAE-O), whilst we use Kmean and silhouette values [R29] to find the reasonable distributions of the benign data samples for the second case. 
 
    As a result, CTVAE combined with Kmean (CTVAE-K) is trained on new multi-class datasets instead of the original binary datasets. As can be seen,  the performance of CTVAE-K is better than that of CTVAE-O. For example, the accuracy and Fscore achieved by CTVAE-O on IDS2017 are $ 0.979 $, and $ 0.979 $, respectively, which are lower than those by CTVAE-K, $ 0.981 $ and $ 0.982 $, respectively. The results show that if the number of distributions in the training set is reasonably chosen ($C$ in Algorithm 1), it can make the training set less imbalanced and then boost the performance of CTVAE in terms of accuracy and Fscore. This shows the superior performance of CTVAE when addressing the problem of imbalanced data.  

\subsection{Comparison with Supervised Dimensionality Reduction Technique}

 \begin{table}[t]
		\centering
		\caption{{CTVAE vs. LDA \cite{R22, R4}.}}
		\label{tab:ctvae_vs_lda}
		\begin{tabular}{|c|cc|cc|}
			\hline
			\multirow{2}{*}{\textbf{Datasets}} & \multicolumn{2}{c|}{\textbf{Accuracy}}  & \multicolumn{2}{c|}{\textbf{Fscore}}\\ \cline{2-5} 
			& \multicolumn{1}{c|}{\textbf{LDA}}    & \textbf{CTVAE}   & \multicolumn{1}{c|}{\textbf{LDA}}    & \textbf{CTVAE}   \\ \hline
			IoT-01& \multicolumn{1}{c|}{92.4} & \textbf{93.0}  & \multicolumn{1}{c|}{89.7} & \textbf{90.9}  \\ \hline
			IoT-02& \multicolumn{1}{c|}{94.4} & \textbf{95.0}  & \multicolumn{1}{c|}{92.3} & \textbf{93.5}  \\ \hline
			IoT-03& \multicolumn{1}{c|}{93.2} & \textbf{93.9}  & \multicolumn{1}{c|}{90.8} & \textbf{92.2}  \\ \hline
			IoT-04& \multicolumn{1}{c|}{90.9} & \textbf{91.8}  & \multicolumn{1}{c|}{87.7} & \textbf{89.5}  \\ \hline
			IoT-05& \multicolumn{1}{c|}{92.2} & \textbf{93.1}  & \multicolumn{1}{c|}{89.4} & \textbf{91.1}  \\ \hline
			IoT-06& \multicolumn{1}{c|}{\textbf{94.6}} & 94.5& \multicolumn{1}{c|}{\textbf{93.4}} & 93.3\\ \hline
			IoT-07& \multicolumn{1}{c|}{\textbf{96.7}} & 96.6& \multicolumn{1}{c|}{\textbf{95.8}} & 95.5\\ \hline
			IoT-08& \multicolumn{1}{c|}{94.8} & \textbf{100.0} & \multicolumn{1}{c|}{95.9} & \textbf{100.0} \\ \hline
			IoT-09& \multicolumn{1}{c|}{75.1} & \textbf{98.9}  & \multicolumn{1}{c|}{82.2} & \textbf{99.0}  \\ \hline
			IoT-10& \multicolumn{1}{c|}{72.7} & \textbf{99.7}  & \multicolumn{1}{c|}{80.7} & \textbf{99.8}  \\ \hline
			IoT-11& \multicolumn{1}{c|}{75.2} & \textbf{99.7}  & \multicolumn{1}{c|}{80.6} & \textbf{99.7}  \\ \hline
		\end{tabular}
	\end{table}

As a performance benchmark for CTVAE, we adopt the Linear Discriminant Analysis (LDA) \cite{R22, R4}, the most famous technique for dimensionality reduction using supervised learning for IDS. The comparison is shown in Table \ref{tab:ctvae_vs_lda}. LDA aims to maximize the ratio of the between-class variance to the within-class variance to separate data samples of different classes \cite{R22, R4}. The RF classifier is used to evaluate the performance of representation data of CTVAE and LDA.
		As can be seen, the performance of CTVAE is better than that of LDA on 11 IoT datasets. The accuracy and Fscore obtained by CTVAE are significantly greater than those of LDA for the first five datasets, from IoT-01 to IoT-05, while the performance of LDA is slightly better than that of CTVAE on two datasets, i.e., IoT-06 and IoT-07. For example, the accuracy and Fscore obtained by CTVAE on IoT-04 are $91.8\%$ and $89.5\%$ compared to $90.9\%$ and $87.7\%$ of LDA, respectively. Remarkably, the results of CTVAE show superior performance over four binary datasets, i.e. from IoT-08 to IoT-11. This is because LDA can only transform data input to 1-dimensional space with binary datasets \footnote{The number of dimensions obtained by LDA for dimensionality reduction is $d_\textbf{z}$ and $d_\textbf{z} \leq \min((|C|-1), d_{input})$ where $|C|$ is the number of classes and $d_{input}$ is the number of dimensions of the input data \cite{R22}.} \cite{R22, R4}, whilst CTVAE can project data from the input space (115-dimensional space) into 10-dimensional space. This decreases the accuracy and Fscore of LDA on all binary datasets.  
 
\subsection{Impact of Transforming the Mean of Distributions}

\begin{table}[t]
	\centering
	\setlength\tabcolsep{3.0pt}
	\caption{Comparing CTVAE-transform with CTVAE-fix. }
	\label{tab:tab_multi_CTVAE_fixed}
	\begin{tabular}{|c|cc|cc|}
		\hline
		\multirow{3}{*}{\textbf{Datasets}} & \multicolumn{2}{c|}{\textbf{Accuracy}}& \multicolumn{2}{c|}{\textbf{Fscore}} \\ \cline{2-5} 
		& \multicolumn{1}{c|}{\textbf{CTVAE}}   & \textbf{CTVAE}   & \multicolumn{1}{c|}{\textbf{CTVAE}}   & \textbf{CTVAE}   \\ \cline{2-5} 
		& \multicolumn{1}{c|}{\textbf{fix}}   & \textbf{transform}  & \multicolumn{1}{c|}{\textbf{fix}}   & \textbf{transform}  \\ \hline
		IoT-01& \multicolumn{1}{c|}{92.0}& \textbf{93.0}  & \multicolumn{1}{c|}{89.1}& \textbf{90.9}  \\ \hline
		IoT-02& \multicolumn{1}{c|}{94.4}& \textbf{95.0}  & \multicolumn{1}{c|}{92.6}& \textbf{93.5}  \\ \hline
		IoT-03& \multicolumn{1}{c|}{93.7}& \textbf{93.9}  & \multicolumn{1}{c|}{91.9}& \textbf{92.2}  \\ \hline
		IoT-04& \multicolumn{1}{c|}{90.1}& \textbf{91.8}  & \multicolumn{1}{c|}{87.2}& \textbf{89.5}  \\ \hline
		IoT-05& \multicolumn{1}{c|}{91.9}& \textbf{93.1}  & \multicolumn{1}{c|}{88.8}& \textbf{91.1}  \\ \hline
		IoT-06& \multicolumn{1}{c|}{84.6}& \textbf{94.5}  & \multicolumn{1}{c|}{83.8}& \textbf{93.3}  \\ \hline
		IoT-07& \multicolumn{1}{c|}{94.6}& \textbf{96.6}  & \multicolumn{1}{c|}{93.7}& \textbf{95.5}  \\ \hline
		IoT-08& \multicolumn{1}{c|}{98.3}& \textbf{100.0} & \multicolumn{1}{c|}{98.4}& \textbf{100.0} \\ \hline
		IoT-09& \multicolumn{1}{c|}{98.0}& \textbf{98.9}  & \multicolumn{1}{c|}{98.1}& \textbf{99.0}  \\ \hline
	\end{tabular}
	
\end{table}

\begin{table*}[htp]
	\renewcommand{\arraystretch}{1}
	\scriptsize\addtolength{\tabcolsep}{-6pt}
	\begin{center}
		\begin{tabular}{m{1em}ccc}		
			&\begin{subfigure}{0.31\textwidth}\centering\includegraphics[width=\linewidth]{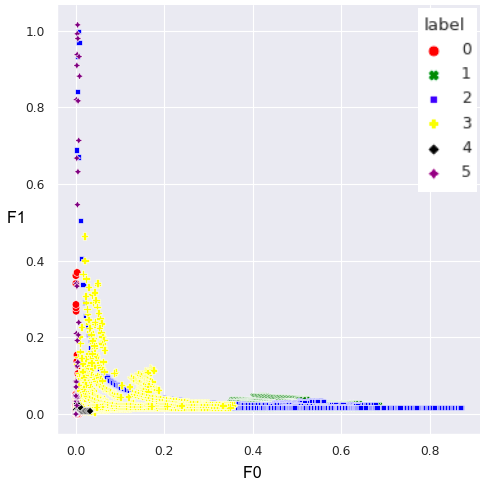}\end{subfigure}
			&\begin{subfigure}{0.31\textwidth}\centering\includegraphics[width=\linewidth]{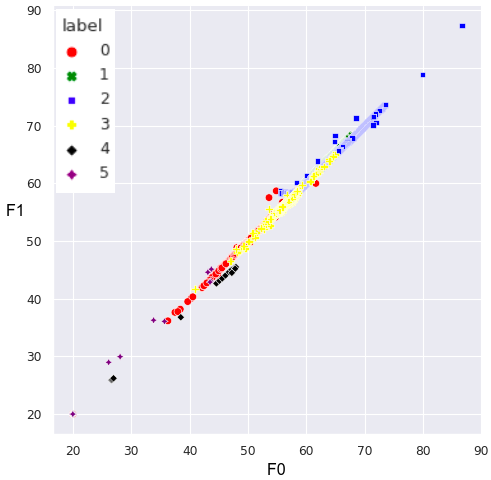}\end{subfigure}
			&\begin{subfigure}{0.31\textwidth}\centering\includegraphics[width=\linewidth]{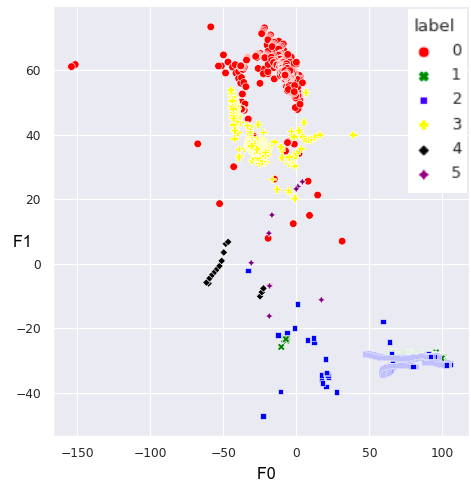}\end{subfigure}\\[0ex]
			&a. Original dataset& b. CTVAE-fix &c. CTVAE-transform\\ 	
		\end{tabular}
	\end{center}
	\captionof{figure}{Visualizing the original data and the reconstruction representation CTVAE-fix and CTVAE-transform on IoT-01.}
	\label{fig:data_rep_fixed_scaled}
\end{table*}	

In this subsection, we investigate the role of transforming the mean of the distribution of each class in the latent space of CTVAE. In other words, we compare two techniques to calculate the mean of the distribution. In the first technique, $\hat{\mu}^{(c)}_j$ is fixed at $S \times c$ where $S$ is a hyper-parameter, $c \in C$ is the label of each class~\footnote{The value of $S$ is set at 20 in this experiment.}. This way of calculating the mean is similar to that used in MVAE~\cite{R1}, however, it is extended from 2 classes in ~\cite{R1} to multiple classes in this paper. This technique is named as CTVAE-fix. In the second technique, $\hat{\mu}^{(c)}_j$ is calculated using Algorithm \ref{algo:separatedZAlgo} in this paper. This technique is named as CTVAE-transform. 


Table \ref{tab:tab_multi_CTVAE_fixed} shows the performance of CTVAE-fix and CTVAE-transform. As can be observed, CTVAE-transform is better than CTVAE-fix on all tested datasets. For example, on IoT-08, the accuracy and Fscore of CTVAE-fix are 98.3\% and 98.4\%, respectively, while both values of CTVAE-transform are 100\%. Similar results are also observed in the other datasets. The results show the effectiveness of using Algorithm \ref{algo:separatedZAlgo} to transform the mean of the distribution of each class to separate them in the latent space. 


We also visualize the reconstruction representation of CTVAE-fix and CTVAE-transform on IoT-01 in Fig. \ref{fig:data_rep_fixed_scaled}. In this figure, the two first features of the original data and the reconstruction representation are used for the visualization. It is apparent that the data samples of different classes are overlapped in the original data (Fig. \ref{fig:data_rep_fixed_scaled}a). Thus, machine learning models trained directly on the original data may not perform well (evidenced by the result in Table~\ref{tab:tab_multi_label_TVAE_vs_ALL}). Using CTVAE-fix, the original data is transformed to become the reconstruction representation that is mostly lined on a hyperplane (Fig. \ref{fig:data_rep_fixed_scaled}b). However, the data samples of the reconstruction representation of CTVAE-fix are still mixed. Thus, the performance of machine learning models trained on the reconstruction representation of CTVAE-fix is also not satisfactory. Contrary to the original data and the reconstruction representation of CTVAE-fix, the reconstruction representation of CTVAE-transform is far more separable. The data samples of different classes in Fig. \ref{fig:data_rep_fixed_scaled}c are stretched out to different areas in the space. Thus the machine learning models can achieve more convincing results using this representation. 


\subsection{Model Complexity}
\label{l_complexity}
\begin{table}[htp]
		\centering
		\setlength\tabcolsep{0.2pt}
		\caption{{Model complexity of CTVAE compared to MAE and MVAE on IoT-02 dataset.}}
		\label{tab:tab_complexity_vs_baseline}
		\begin{tabular}{|c|c|c|c|c|}
			\hline
			\textbf{Models} & \begin{tabular}[c]{@{}c@{}}Model size\\  In hard disk\\  (KB)\end{tabular} & \begin{tabular}[c]{@{}c@{}}Avg running\\ time for extracting\\ a data sample \\ (seconds)\end{tabular} & \begin{tabular}[c]{@{}c@{}}Avg running\\ time for training\\ an epoch \\ (seconds)\end{tabular} &  \\ \hline
			MAE  & {108.9}   & 1.66E-06  & {8.24}  &  \\ \hline
			MVAE & 425.2 & 1.89E-06  & 11.00   &  \\ \hline
			CTVAE& 617.7 & \textbf{1.62E-06}& 12.45   &  \\ \hline
		\end{tabular}
	\end{table}	
	
	\begin{table}[htp]
		\centering
		\setlength\tabcolsep{0.2pt}
		\caption{{Model complexity of CTVAE compared to Xgb on IoT-02 dataset.}}
		\label{tab:tab_ctvae_vs_xgb13}
		\begin{tabular}{|c|c|c|}
			\hline
			\textbf{Criteria} & \textbf{CTVAE + RF} & \textbf{Xgb}    \\ \hline
			avg time (seconds) & \textbf{1.72E-06}   & 1.54E-05   \\ \hline
			Model size in hard disk (KB) & 618.33   & \textbf{549.04} \\ \hline
		\end{tabular}
	\end{table}

To show the effectiveness of the CTVAE in terms of running time for extracting data and model size in hard disk, we have compared the CTVAE with those of MAE and MVAE that are applied to IoT devices \cite{R1}, as observed in Table \ref{tab:tab_complexity_vs_baseline}. We have also compared the full process of an IDS when combining the CTVAE with a classifier, i.e., Random Forests (RF), to Xgb classifier, as observed in Table \ref{tab:tab_ctvae_vs_xgb13}.
		We measure three metrics including model size in hard disk (model-size), average running time (time-extraction) for extracting a data sample to obtain representation data ($\hat{\textbf{z}}$ for CTVAE and $\textbf{z}$ for MAE and MVAE), and average running time for training an epoch (time-epoch). To calculate time-extraction, and time-epoch for three models, we use the IoT-02 dataset for testing. To measure the model size, we use the $pickle$ $tf.save$ function of Tensorflow to save three neuron network models. 
		
		As observed from Table \ref{tab:tab_complexity_vs_baseline}, three models, i.e., MAE, MVAE, and CTVAE consume a low amount of resources in the hard disk. In addition, the time-extraction and time-epoch for the three models are approximately the same, and they are suitable for deploying on IoT devices. This is because the CTVAE only uses the Decoder to extract the data representation instead of using three parts, e.g., an Encoder, a Hermaphrodite, and a Decoder. For example, the model size of three models is lower than $1$ MB which is $108.9$ KB, $425.2$ KB, and $617.7$ KB of MAE, MVAE, and CTVAE, respectively. The time-extraction of CTVAE is $\mbox{1.62E-6}$ seconds, while that of MVAE is $\mbox{1.89E-6}$ seconds. CTVAE spends $12.45$ seconds for each epoch during the training process, whilst the time-epoch of MAE is $8.24$ seconds.
		
		We analyze the usability of CTVAE on IoT devices by comparing the model complexity of CTVAE combined with the RF classifier to Xgb, as observed in Table \ref{tab:tab_ctvae_vs_xgb13}. Note that Xgb is considered to be the state-of-the-art method \cite{bhati2021Xgb, chen2016xgb} that achieves a very good result on a wide range of problems including IoT attack detection. The average running time of CTVAE-RF is faster than that of Xgb ($\mbox{1.72E-6}$ seconds vs $\mbox{1.54E-5}$ seconds, respectively). The model size of the two models is lower than $1$ MB which is suitable for deploying on IoT devices. The results in this subsection show that CTVAE is more relevant to deploy on IoT devices compared to Xgb. This is very important since the resource of an IoT device is often limited, and a large and highly complex model is often not applicable to IoT devices. 

\subsection{CTVAE Simulation}

\begin{table*}[htp]
	\renewcommand{\arraystretch}{1.0}
	\scriptsize\addtolength{\tabcolsep}{-2pt}
	\centering
	\begin{center}
		\begin{tabular}{m{1em}cccc} 				
			&\begin{subfigure}{0.22\textwidth}\centering\includegraphics[width=\linewidth]{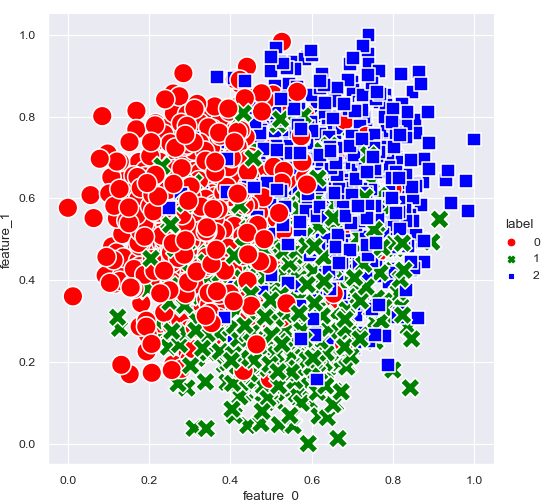}\end{subfigure}
			&\begin{subfigure}{0.22\textwidth}\centering\includegraphics[width=\linewidth]{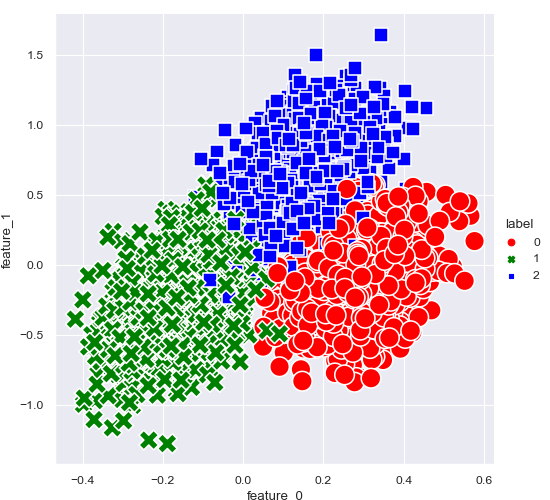}\end{subfigure}
			&\begin{subfigure}{0.22\textwidth}\centering\includegraphics[width=\linewidth]{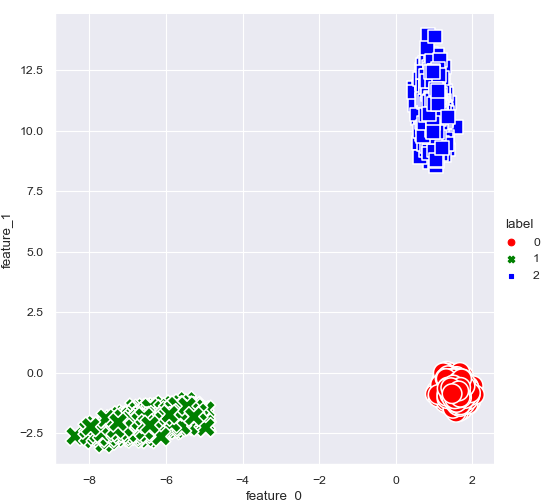}\end{subfigure}
			&\begin{subfigure}{0.22\textwidth}\centering\includegraphics[width=\linewidth]{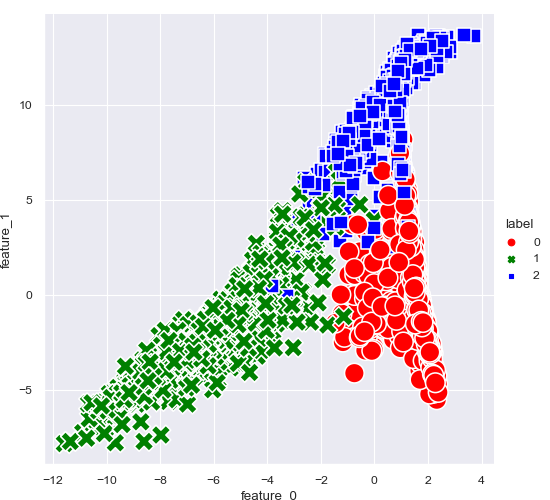}\end{subfigure} \\
            &a. Original data $\textbf{x}$ &b. Mean $\boldsymbol{\mu}$
			& c. Latent Representation $\textbf{z}$ &d.  Reconstruction Representation $\hat{\textbf{z}}$\\
		\end{tabular}
		\captionof{figure}{Simulation of CTVAE.}
		\label{fig:dataRepresentionCTVAE}
	\end{center}
\end{table*}

\begin{table}[htp]
	\renewcommand{\arraystretch}{1.0}
	\scriptsize\addtolength{\tabcolsep}{-4pt}
	\begin{center}
		\begin{tabular}{m{1em}cc}	
			&\begin{subfigure}{0.22\textwidth}\centering\includegraphics[width=\linewidth]{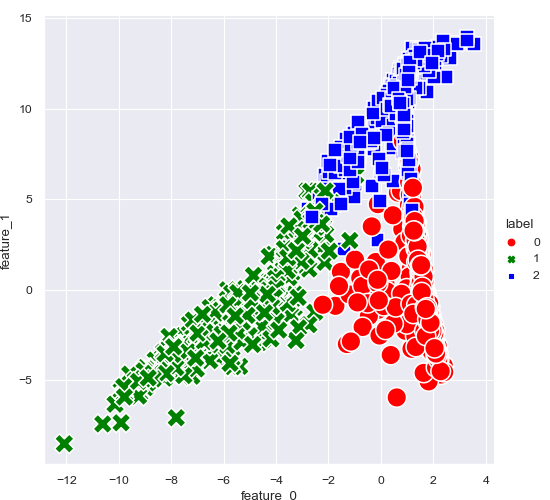}\end{subfigure} 
			
			&\begin{subfigure}{0.22\textwidth}\centering\includegraphics[width=\linewidth]{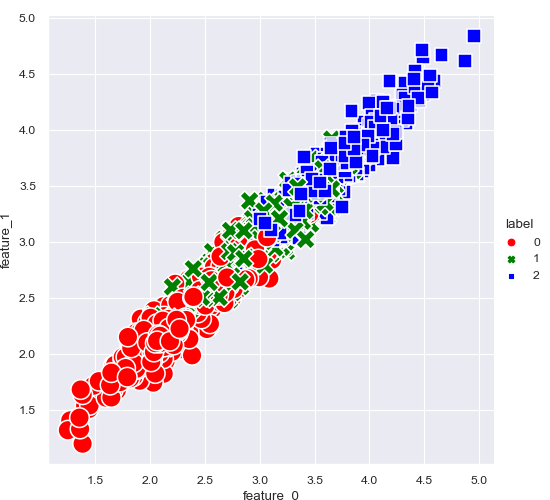}\end{subfigure}

			\\ 
            &a. CTVAE& b. MAE\\ 
		\end{tabular}
		\captionof{figure}{Comparing between CTVAE and MAE.}
		\label{fig:CTVAE_compared_to_MAE}
	\end{center}
\end{table}

In this subsection, we conduct a simulation on an artificial dataset to analyze the behavior of CTVAE. We  create a small artificial dataset including three classes in which the data samples are overlapped. We  use the $make\_blobs$ function in Sklearn~\cite{MakeBlobs} to create the dataset of three  Gaussian distributions. The standard deviation of three Gaussian distributions is set at 0.2. The range of parameter $center\_box$ in \cite{MakeBlobs} is shrunk from 0 to 1, making three Gaussian distributions mixed together. The number of data samples in the training/testing is 3500 and 1500, respectively. The number of features for each data sample is 10. The number of neurons at the latent space $\textbf{z}$ and at the output layer $\hat{\textbf{z}}$ of CTVAE in this experiment is set at 2 to facilitate the visualization.

The result of the simulation of the training process of the CTVAE is presented in Fig. \ref{fig:dataRepresentionCTVAE}. In particular, Fig. \ref{fig:dataRepresentionCTVAE}a shows original datasets using the first two dimensions. Fig. \ref{fig:dataRepresentionCTVAE}b shows the mean $\boldsymbol{\mu}$ of the encoder. Fig. \ref{fig:dataRepresentionCTVAE}c shows the latent space $\textbf{z}$ of CTVAE. Fig. \ref{fig:dataRepresentionCTVAE}d shows reconstruction presentation $\hat{\textbf{z}}$ when the decoder is fed by the original data $\textbf{x}$. First, it can be seen that the original data samples are mixed and they are difficult to classify. Second, the data representation at the output of the encoder, i.e., $\boldsymbol{\mu}$, is still overlapped, as observed in Fig. \ref{fig:dataRepresentionCTVAE}b. However, the data representation at latent space, i.e., $\textbf{z}$, is well separated among three classes. This is achieved by sampling $\textbf{z}$ from the separated Gaussian distribution with different mean and standard deviations based on label information. As a result, the data samples of the reconstruction representation, i.e., $\hat{\textbf{z}}$, of a class are mostly separated from data samples of other classes. This result shows the effectiveness of the CTVAE in transforming an overlapped dataset at the input into a distinguishable dataset at the output.

The last result in this subsection is the comparison of the representation of CTVAE and MAE on the testing set of the artificial dataset. Fig. \ref{fig:CTVAE_compared_to_MAE} presents reconstruction representation, i.e., $\hat{\textbf{z}}$ of CTVAE and the latent representation of MAE. In general, the representation of CTVAE (Fig. \ref{fig:CTVAE_compared_to_MAE}a) is much more distinguishable than the representation of MAE (Fig. \ref{fig:CTVAE_compared_to_MAE}b). The reason is that MAE attempts to constrain its representation to the predefined centers and thus it is harder to train for a dataset of multiple classes and with the small number of data samples \cite{R1}. Conversely, CTVAE transforms the mean of the distribution to make the more separate. Thus CTVAE is easier to train to separate the classes in the dataset.

\subsection{Representation Quality}
	\begin{table}[ht]
		\centering
		\caption{{  Compare the quality of original data with representation data of CTVAE.}}
		\label{tab:data_qualify_compared12}
		\begin{tabular}{|c|cc|cc|}
			\hline
			\multirow{2}{*}{\textbf{Datasets}} & \multicolumn{2}{c|}{\textbf{$d_{bet}$}}& \multicolumn{2}{c|}{\textbf{$d_{wit}$}} \\ \cline{2-5} 
			& \multicolumn{1}{c|}{\textbf{Original}} & \textbf{CTVAE} & \multicolumn{1}{c|}{\textbf{Original}} & \textbf{CTVAE} \\ \hline
			IoT-01& \multicolumn{1}{c|}{\textbf{1.15}}& 0.89& \multicolumn{1}{c|}{1.65E-03} & \textbf{3.29E-04}  \\ \hline
			IoT-02& \multicolumn{1}{c|}{1.15}   & \textbf{1.59}& \multicolumn{1}{c|}{2.08E-03} & \textbf{5.25E-04}  \\ \hline
			IoT-03& \multicolumn{1}{c|}{1.16}   & \textbf{2.69}& \multicolumn{1}{c|}{4.04E-03} & \textbf{6.22E-04}  \\ \hline
			IoT-04& \multicolumn{1}{c|}{1.24}   & \textbf{2.54}& \multicolumn{1}{c|}{1.92E-03} & \textbf{2.29E-04}  \\ \hline
			IoT-05& \multicolumn{1}{c|}{1.17}   & \textbf{2.23}& \multicolumn{1}{c|}{2.90E-03} & \textbf{5.25E-04}  \\ \hline
			IoT-06& \multicolumn{1}{c|}{\textbf{4.12}}& 2.65& \multicolumn{1}{c|}{1.38E-02} & \textbf{9.11E-04}  \\ \hline
			IoT-07& \multicolumn{1}{c|}{3.20}   & \textbf{4.18}& \multicolumn{1}{c|}{9.76E-03} & \textbf{3.07E-03}  \\ \hline
		\end{tabular}
	\end{table}

    Table \ref{tab:data_qualify_compared12} shows the comparison of the quality of data between the seven original IoT datasets and their data representation extracted from the CTVAE by using two measures, i.e., $d_{bet}$ and $d_{wit}$. As can be seen, the values of $d_{bet}$ of the representation data are mostly greater than those of the original data. In addition, the values of $d_{wit}$ of the representation data are significantly smaller than those of the original data. These can show the fact that the means of different classes of the data representation of CTVAE are far from each other, whilst data samples of a class are forced to its mean. Therefore, the quality of the data representation of CTVAE is better than that of the original dataset.
 
\section{Conclusion and Future Work}
\label{conclusion}
In this paper, we proposed a novel deep neural network model for transforming the original dataset into a more distinguishable dataset. The proposed model was called CTVAE (Constrained Twin Variational Auto-Encoder). CTVAE was trained in a supervised manner to transform the latent representation into a more stable and separable reconstruction representation. The CTVAE could address the challenge of IDSs with complex data, i.e., high dimension and heterogeneous data formats. The CTVAE also enhanced and facilitated the IDS classifiers in the detection of anomalies by presenting/feeding them with more separable/distinguishable data samples. In addition, CTVAE combined with K-Mean and silhouette values to introduce a solution to address the imbalanced data in IoT IDS systems. Finally, CTVAE required less storage/memory and computing power and hence can be more suitable for IoT IDS systems. We evaluated the performance of CTVAE using extensive experiments on 11 IoT botnet datasets. The effectiveness of CTVAE was measured using a popular classifier, i.e., RF. The results were compared with some baseline methods and the state-of-the-art techniques for RL, i.e., MAE. Overall, CTVAE was better than recent RL models, i.e., MAE, MVAE, CSAEC, and popular machine learning models, i.e., RF, SVM, DT, and RF, and a state-of-the-art machine learning model, i.e., Xgb. The simulation results on an artificial dataset also revealed that CTVAE achieved its objective by transforming the overlapped dataset into a well-separated dataset. The simulation results also provided an explanation for the superior performance of CTVAE to the others.  

In the future, one can extend our work in different directions. First, the proposed models in this article were only examined on the IoT attack datasets. It is also interesting to extend these models and test them on cyberattack datasets, and other fields such as computer vision and natural language processing. Second, the hyper-parameters in the loss function of CTVAE, i.e., $\beta_1$, $\beta_2$, $\beta_3$, and $\beta_4$, were determined through trial and error. It could be better to find an approach to automatically select good values for each dataset. Last but not least, the twin models can also be used to learn the latent representation of other generative models based on VAE.

\section*{Acknowledgment}
This research was funded by Vingroup Innovation Foundation (VINIF) under project code VINIF.2023.DA059
\ifCLASSOPTIONcaptionsoff
\newpage
\fi
\vspace{0mm}
\bibliographystyle{IEEEtran}
\bibliography{IEEEabrv,library}{}

\begin{thebibliography}{10}
\providecommand{\url}[1]{#1}
\csname url@samestyle\endcsname
\providecommand{\newblock}{\relax}
\providecommand{\bibinfo}[2]{#2}
\providecommand{\BIBentrySTDinterwordspacing}{\spaceskip=0pt\relax}
\providecommand{\BIBentryALTinterwordstretchfactor}{4}
\providecommand{\BIBentryALTinterwordspacing}{\spaceskip=\fontdimen2\font plus
\BIBentryALTinterwordstretchfactor\fontdimen3\font minus
  \fontdimen4\font\relax}
\providecommand{\BIBforeignlanguage}[2]{{%
\expandafter\ifx\csname l@#1\endcsname\relax
\typeout{** WARNING: IEEEtran.bst: No hyphenation pattern has been}%
\typeout{** loaded for the language `#1'. Using the pattern for}%
\typeout{** the default language instead.}%
\else
\language=\csname l@#1\endcsname
\fi
#2}}
\providecommand{\BIBdecl}{\relax}
\BIBdecl

\bibitem{R11}
L.~Yang, A.~Moubayed, and A.~Shami, ``Mth-ids: A multitiered hybrid intrusion
  detection system for internet of vehicles,'' \emph{IEEE Internet of Things
  Journal}, vol.~9, no.~1, pp. 616--632, Jan. 2021.

\bibitem{R7}
S.~I. Popoola, B.~Adebisi, M.~Hammoudeh, G.~Gui, and H.~Gacanin, ``Hybrid deep
  learning for botnet attack detection in the internet-of-things networks,''
  \emph{IEEE Internet of Things Journal}, vol.~8, no.~6, pp. 4944--4956, Oct.
  2020.

\bibitem{R4}
P.~Xanthopoulos, P.~M. Pardalos, T.~B. Trafalis, P.~Xanthopoulos, P.~M.
  Pardalos, and T.~B. Trafalis, ``Linear discriminant analysis,'' \emph{Robust
  data mining}, pp. 27--33, Jan. 2013.

\bibitem{R8}
I.~Ullah and Q.~H. Mahmoud, ``Design and development of a deep learning-based
  model for anomaly detection in iot networks,'' \emph{IEEE Access}, vol.~9,
  pp. 103\,906--103\,926, Jul. 2021.

\bibitem{R9}
G.~Abdelmoumin, D.~B. Rawat, and A.~Rahman, ``On the performance of machine
  learning models for anomaly-based intelligent intrusion detection systems for
  the internet of things,'' \emph{IEEE Internet of Things Journal}, vol.~9,
  no.~6, pp. 4280--4290, Aug. 2021.

\bibitem{hajiheidari2019intrusion}
S.~Hajiheidari, K.~Wakil, M.~Badri, and N.~J. Navimipour, ``Intrusion detection
  systems in the internet of things: A comprehensive investigation,''
  \emph{Computer Networks}, vol. 160, pp. 165--191, Sept. 2019.

\bibitem{ids2019survey}
A.~Khraisat, I.~Gondal, P.~Vamplew, and J.~Kamruzzaman, ``Survey of intrusion
  detection systems: techniques, datasets and challenges,''
  \emph{Cybersecurity}, vol.~2, no.~1, pp. 1--22, Jul. 2019.

\bibitem{ferrag2020deep}
M.~A. Ferrag, L.~Maglaras, S.~Moschoyiannis, and H.~Janicke, ``Deep learning
  for cyber security intrusion detection: Approaches, datasets, and comparative
  study,'' \emph{Journal of Information Security and Applications}, vol.~50,
  no.~1, p. 102419, 2020.

\bibitem{bengio2013rep}
Y.~Bengio, A.~Courville, and P.~Vincent, ``Representation learning: A review
  and new perspectives,'' \emph{IEEE Transactions on Pattern Analysis and
  Machine Intelligence}, vol.~35, no.~8, pp. 1798--1828, Mar. 2013.

\bibitem{Goodfellow}
I.~Goodfellow, Y.~Bengio, and A.~Courville, \emph{Deep Learning}.\hskip 1em
  plus 0.5em minus 0.4em\relax The MIT Press, 2016.

\bibitem{R26}
Y.~Bengio, A.~Courville, and P.~Vincent, ``Representation learning: A review
  and new perspectives,'' \emph{IEEE Transactions on Pattern Analysis and
  Machine Intelligence}, vol.~35, no.~8, pp. 1798--1828, Mar. 2013.

\bibitem{dao2021stacked}
T.-N. Dao and H.~Lee, ``Stacked autoencoder-based probabilistic feature
  extraction for on-device network intrusion detection,'' \emph{IEEE Internet
  of Things Journal}, vol.~9, no.~16, pp. 14\,438--14\,451, Aug. 2022.

\bibitem{R27}
J.~Sun, X.~Wang, N.~Xiong, and J.~Shao, ``Learning sparse representation with
  variational auto-encoder for anomaly detection,'' \emph{IEEE Access}, vol.~6,
  pp. 33\,353--33\,361, Jun. 2018.

\bibitem{R1}
L.~{Vu}, V.~L. {Cao}, Q.~U. {Nguyen}, D.~N. {Nguyen}, D.~T. {Hoang}, and
  E.~{Dutkiewicz}, ``Learning latent representation for iot anomaly
  detection,'' \emph{IEEE Transactions on Cybernetics}, pp. 1--14, Sept. 2020.

\bibitem{TomczakW17}
J.~Tomczak and M.~Welling, ``Vae with a vampprior,'' in \emph{International
  Conference on Artificial Intelligence and Statistics}.\hskip 1em plus 0.5em
  minus 0.4em\relax Playa Blanca, Lanzarote, Canary Islands: PMLR, 2018, pp.
  1214--1223.

\bibitem{R20}
P.~V. Dinh, N.~Q. Uy, D.~N. Nguyen, D.~T. Hoang, S.~P. Bao, and E.~Dutkiewicz,
  ``Twin variational auto-encoder for representation learning in iot intrusion
  detection,'' in \emph{2022 IEEE Wireless Communications and Networking
  Conference (WCNC)}.\hskip 1em plus 0.5em minus 0.4em\relax Austin, TX, USA:
  IEEE, 2022, pp. 848--853.

\bibitem{R29}
R.~Llet{\i}, M.~C. Ortiz, L.~A. Sarabia, and M.~S. S{\'a}nchez, ``Selecting
  variables for k-means cluster analysis by using a genetic algorithm that
  optimises the silhouettes,'' \emph{Analytica Chimica Acta}, vol. 515, no.~1,
  pp. 87--100, Jul. 2004.

\bibitem{R13}
Y.~Meidan, M.~Bohadana, Y.~Mathov, Y.~Mirsky, A.~Shabtai, D.~Breitenbacher, and
  Y.~Elovici, ``N-baiot—network-based detection of iot botnet attacks using
  deep autoencoders,'' \emph{IEEE Pervasive Computing}, vol.~17, no.~3, pp.
  12--22, Mar. 2018.

\bibitem{R10}
C.~Yin, S.~Zhang, J.~Wang, and N.~N. Xiong, ``Anomaly detection based on
  convolutional recurrent autoencoder for iot time series,'' \emph{IEEE
  Transactions on Systems, Man, and Cybernetics: Systems}, vol.~52, no.~1, pp.
  112--122, Jan. 2020.

\bibitem{R16}
T.-N. Dao and H.~Lee, ``Stacked autoencoder-based probabilistic feature
  extraction for on-device network intrusion detection,'' \emph{IEEE Internet
  of Things Journal}, vol.~9, no.~16, pp. 14\,438--14\,451, May 2021.

\bibitem{R18}
W.~Luo, J.~Li, J.~Yang, W.~Xu, and J.~Zhang, ``Convolutional sparse
  autoencoders for image classification,'' \emph{IEEE Transactions on Neural
  Networks and Learning Systems}, vol.~29, no.~7, pp. 3289--3294, Jul. 2018.

\bibitem{R19}
M.~Al-Qatf, Y.~Lasheng, M.~Al-Habib, and K.~Al-Sabahi, ``Deep learning approach
  combining sparse autoencoder with svm for network intrusion detection,''
  \emph{IEEE Access}, vol.~6, pp. 52\,843--52\,856, Sept. 2018.

\bibitem{R23}
T.~Chen and C.~Guestrin, ``Xgboost: A scalable tree boosting system,'' in
  \emph{Proceedings of the 22nd ACM SIGKDD International Conference on
  Knowledge Discovery and Data Mining}, San Francisco California USA, 2016, pp.
  785--794.

\bibitem{kleinbaum2002logistic}
D.~G. Kleinbaum, K.~Dietz, M.~Gail, M.~Klein, and M.~Klein, \emph{Logistic
  Regression}.\hskip 1em plus 0.5em minus 0.4em\relax Springer, 2002.

\bibitem{LSVM}
\BIBentryALTinterwordspacing
``Linear support vector machine.'' [Online]. Available:
  \url{https://scikit-learn.org/stable/modules/generated/sklearn.svm.LinearSVC.html.}
\BIBentrySTDinterwordspacing

\bibitem{DT}
\BIBentryALTinterwordspacing
``Decision tree.'' [Online]. Available:
  \url{https://scikit-learn.org/stable/modules/generated/sklearn.tree.DecisionTreeClassifier.html.}
\BIBentrySTDinterwordspacing

\bibitem{RF}
\BIBentryALTinterwordspacing
``Random forest.'' [Online]. Available:
  \url{https://scikit-learn.org/stable/modules/generated/sklearn.ensemble.RandomForestClassifier.html.}
\BIBentrySTDinterwordspacing

\bibitem{sindhu2012DT}
S.~S.~S. Sindhu, S.~Geetha, and A.~Kannan, ``Decision tree based light weight
  intrusion detection using a wrapper approach,'' \emph{Expert Systems with
  applications}, vol.~39, no.~1, pp. 129--141, Jan. 2012.

\bibitem{moon2017DT}
D.~Moon, H.~Im, I.~Kim, and J.~H. Park, ``Dtb-ids: an intrusion detection
  system based on decision tree using behavior analysis for preventing apt
  attacks,'' \emph{The Journal of supercomputing}, vol.~73, no.~7, pp.
  2881--2895, Dec. 2017.

\bibitem{chen2009SVM}
R.-C. Chen, K.-F. Cheng, Y.-H. Chen, and C.-F. Hsieh, ``Using rough set and
  support vector machine for network intrusion detection system,'' in
  \emph{2009 First Asian Conference on Intelligent Information and Database
  Systems}.\hskip 1em plus 0.5em minus 0.4em\relax Dong hoi, Vietnam: IEEE,
  2009, pp. 465--470.

\bibitem{hasan2014SVM}
M.~A.~M. Hasan, M.~Nasser, B.~Pal, and S.~Ahmad, ``Support vector machine and
  random forest modeling for intrusion detection system (ids),'' \emph{Journal
  of Intelligent Learning Systems and Applications}, vol.~6, no.~1, Aug. 2014.

\bibitem{li2020AERF}
X.~Li, W.~Chen, Q.~Zhang, and L.~Wu, ``Building auto-encoder intrusion
  detection system based on random forest feature selection,'' \emph{Computers
  \& Security}, vol.~95, p. 101851, Aug. 2020.

\bibitem{kim2006RF}
D.~S. Kim, S.~M. Lee, and J.~S. Park, ``Building lightweight intrusion
  detection system based on random forest,'' in \emph{International Symposium
  on Neural Networks}.\hskip 1em plus 0.5em minus 0.4em\relax Chengdu, China:
  Springer, 2006, pp. 224--230.

\bibitem{chen2016xgb}
T.~Chen and C.~Guestrin, ``Xgboost: A scalable tree boosting system,'' in
  \emph{Proceedings of the 22nd ACM SIGKDD International Conference on
  Knowledge Discovery and Data Mining}, New York, NY, United States, 2016, pp.
  785--794.

\bibitem{bhati2021Xgb}
B.~S. Bhati, G.~Chugh, F.~Al-Turjman, and N.~S. Bhati, ``An improved ensemble
  based intrusion detection technique using xgboost,'' \emph{Transactions on
  Emerging Telecommunications Technologies}, vol.~32, no.~6, p. e4076, Aug.
  2021.

\bibitem{vu2020deep}
L.~Vu, Q.~U. Nguyen, D.~N. Nguyen, D.~T. Hoang, and E.~Dutkiewicz, ``Deep
  transfer learning for iot attack detection,'' \emph{IEEE Access}, vol.~8, pp.
  107\,335--107\,344, Jun. 2020.

\bibitem{vincent2010stacked}
P.~Vincent, H.~Larochelle, I.~Lajoie, Y.~Bengio, and P.-A. Manzagol, ``Stacked
  denoising autoencoders: Learning useful representations in a deep network
  with a local denoising criterion,'' \emph{Journal of Machine Learning
  Research}, vol.~11, no.~12, pp. 3371--3408, Dec. 2010.

\bibitem{Shone2018}
N.~{Shone}, T.~N. {Ngoc}, V.~D. {Phai}, and Q.~{Shi}, ``A deep learning
  approach to network intrusion detection,'' \emph{IEEE Transactions on
  Emerging Topics in Computational Intelligence}, vol.~2, no.~1, pp. 41--50,
  Feb. 2018.

\bibitem{hinton2006reducing}
G.~E. Hinton and R.~R. Salakhutdinov, ``Reducing the dimensionality of data
  with neural networks,'' \emph{Science}, vol. 313, no. 5786, pp. 504--507,
  Jul. 2006.

\bibitem{phan2018Tree}
A.~V. Phan, P.~N. Chau, M.~Le~Nguyen, and L.~T. Bui, ``Automatically
  classifying source code using tree-based approaches,'' \emph{Data \&
  Knowledge Engineering}, vol. 114, pp. 12--25, Mar. 2018.

\bibitem{Abdulhammed2018}
R.~Abdulhammed, M.~Faezipour, A.~Abuzneid, and A.~AbuMallouh, ``Deep and
  machine learning approaches for anomaly-based intrusion detection of
  imbalanced network traffic,'' \emph{IEEE Sensors Letters}, vol.~3, no.~1, pp.
  1--4, Nov. 2018.

\bibitem{jang2017}
E.~Jang, S.~Gu, and B.~Poole, ``Categorical reparameterization with
  gumbel-softmax,'' in \emph{International Conference on Learning
  Representations}, Toulon, France, 2017, pp. 1--12.

\bibitem{bVAE}
C.~P. Burgess, I.~Higgins, A.~Pal, L.~Matthey, N.~Watters, G.~Desjardins, and
  A.~Lerchner, ``Understanding disentangling in beta-vae,'' \emph{arXiv
  preprint arXiv:1804.03599}, 2018.

\bibitem{wu2020vector}
H.~Wu and M.~Flierl, ``Vector quantization-based regularization for
  autoencoders,'' in \emph{Proceedings of the AAAI Conference on Artificial
  Intelligence}, vol.~34, no.~04, New York, USA, 2020, pp. 6380--6387.

\bibitem{Kingma_2019}
D.~P. Kingma and M.~Welling, ``An introduction to variational autoencoders,''
  \emph{Foundations and Trends® in Machine Learning}, vol.~12, no.~4, p.
  307–392, Nov. 2019.

\bibitem{An2015VariationalAB}
J.~An and S.~Cho, ``Variational autoencoder based anomaly detection using
  reconstruction probability,'' \emph{Special Lecture on IE}, vol.~2, no.~1,
  pp. 1--18, Dec. 2015.

\bibitem{Bishop2006}
C.~M. Bishop, \emph{Pattern Recognition and Machine Learning (Information
  Science and Statistics)}.\hskip 1em plus 0.5em minus 0.4em\relax Berlin,
  Heidelberg: Springer-Verlag, 2006, vol. 128, no.~9.

\bibitem{dinh2022balanced}
P.~V. Dinh, D.~N. Nguyen, D.~T. Hoang, N.~Q. Uy, S.~P. Bao, and E.~Dutkiewicz,
  ``Balanced twin auto-encoder for iot intrusion detection,'' in \emph{GLOBECOM
  2022-2022 IEEE Global Communications Conference}.\hskip 1em plus 0.5em minus
  0.4em\relax Rio de Janeiro, Brazil: IEEE, 2022, pp. 3387--3392.

\bibitem{R2}
H.~Zou, T.~Hastie, and R.~Tibshirani, ``Sparse principal component analysis,''
  \emph{Journal of computational and graphical statistics}, vol.~15, no.~2, pp.
  265--286, Jan. 2006.

\bibitem{R22}
A.~Tharwat, T.~Gaber, A.~Ibrahim, and A.~E. Hassanien, ``Linear discriminant
  analysis: A detailed tutorial,'' \emph{AI Communications}, vol.~30, no.~2,
  pp. 169--190, May 2017.

\bibitem{IDS2017}
I.~Sharafaldin, A.~H. Lashkari, and A.~A. Ghorbani, ``Toward generating a new
  intrusion detection dataset and intrusion traffic characterization,''
  \emph{ICISSp}, vol.~1, pp. 108--116, Jan. 2018.

\bibitem{UNSW}
N.~{Moustafa} and J.~{Slay}, ``Unsw-nb15: a comprehensive data set for network
  intrusion detection systems (unsw-nb15 network data set),'' in \emph{2015
  Military Communications and Information Systems Conference (MilCIS)},
  Canberra, Australia, 2015, pp. 1--6.

\bibitem{Scikit}
\BIBentryALTinterwordspacing
``Scikit-learn.'' [Online]. Available: \url{https://scikit-learn.org/stable/.}
\BIBentrySTDinterwordspacing

\bibitem{kingma2017adam}
K.~DP and J.~Ba, ``Adam: A method for stochastic optimization,'' in \emph{Proc.
  of the 3rd International Conference for Learning Representations (ICLR)}, San
  Diego, California, US, 2015, pp. 1--15.

\bibitem{glorot10a}
X.~Glorot and Y.~Bengio, ``Understanding the difficulty of training deep
  feedforward neural networks,'' in \emph{Proceedings of the Thirteenth
  International Conference on Artificial Intelligence and Statistics}, vol.~9,
  Chia Laguna Resort, Sardinia, Italy, 2010, pp. 249--256.

\bibitem{R12}
M.~Nicolau, J.~McDermott \emph{et~al.}, ``Learning neural representations for
  network anomaly detection,'' \emph{IEEE Transactions on Cybernetics},
  vol.~49, no.~8, pp. 3074--3087, Jun. 2018.

\bibitem{park2020low}
J.~Park, J.~Lee, and D.~Sim, ``Low-complexity cnn with 1d and 2d filters for
  super-resolution,'' \emph{Journal of Real-Time Image Processing}, vol.~17,
  no.~6, pp. 2065--2076, Jun. 2020.

\bibitem{gridSVM2016}
I.~Syarif, A.~Prugel-Bennett, and G.~Wills, ``Svm parameter optimization using
  grid search and genetic algorithm to improve classification performance,''
  \emph{Telkomnika}, vol.~14, no.~4, p. 1502, Apr. 2016.

\bibitem{gridDT2019}
B.~Shekar and G.~Dagnew, ``Grid search-based hyperparameter tuning and
  classification of microarray cancer data,'' in \emph{2019 Second
  International Conference on Advanced Computational and Communication
  Paradigms (ICACCP)}.\hskip 1em plus 0.5em minus 0.4em\relax Gangtok, Sikkim,
  India: IEEE, 2019, pp. 1--8.

\bibitem{gridRF2019}
P.~Probst, M.~N. Wright, and A.-L. Boulesteix, ``Hyperparameters and tuning
  strategies for random forest,'' \emph{Wiley Interdisciplinary Reviews: Data
  Mining and Knowledge Discovery}, vol.~9, no.~3, p. e1301, Mar. 2019.

\bibitem{MakeBlobs}
\BIBentryALTinterwordspacing
``Make blobs.'' [Online]. Available:
  \url{https://scikit-learn.org/stable/modules/generated/sklearn
  .datasets.make_blobs.html.}
\BIBentrySTDinterwordspacing

\end{thebibliography}


\end{document}